\documentclass[11pt]{article}

\usepackage[preprint]{acl}
\usepackage{times}
\usepackage{latexsym}

\usepackage[T1]{fontenc}
\usepackage[utf8]{inputenc}

\usepackage{microtype}

\usepackage{inconsolata}

\usepackage{graphicx}
\usepackage{amsmath}
\usepackage{amssymb}
\usepackage{multirow}
\usepackage{booktabs}
\usepackage{pdfpages}
\usepackage{listings}
\title{YRC-Bench: Benchmarking Culturally Aware Machine Translation for Historical Heritage via Tri-Level Semantic Bridging}

\author{
  \textbf{Yan Huang\textsuperscript{1}},
  \textbf{Aobo Li\textsuperscript{1}},
  \textbf{Jiayi Wang\textsuperscript{1}},
  \textbf{Huidong Zhu\textsuperscript{1}}
\\
\\
  \textsuperscript{1}Zhengzhou University of Light Industry
\\
  \small{
    \textbf{Correspondence:} \href{mailto:your_email@domain.edu.cn}{332316010963@zzuli.edu.cn}
  }
}
\begin{document}
\maketitle
\begin{abstract}
Translating historical cultural heritage poses profound challenges for machine translation systems due to the acute semantic asymmetry across eras and cultures, where terse classical syntax frequently obscures domain-specific cultural knowledge. 
While Large Language Models have demonstrated remarkable prowess in generic translation benchmarks, they struggle to negotiate Culturally-Specific Items and diachronic shifts in domain-rich historical texts, such as Yellow River cultural records. 
To bridge this critical gap, we introduce YRC-Bench, the first human-curated, multi-parallel benchmark dedicated to historical heritage translation across Classical Chinese, Modern Chinese, and nine target foreign languages. 
YRC-Bench establishes a tri-level parallel corpus incorporating modern explanatory glosses as explicit semantic bridges and operationalizes translation strategies through paired conserved versus substituted CSI variants categorized by cultural translatability. 
Systematic evaluations of nine leading LLMs across bidirectional translation directions reveal that leveraging modern Chinese glosses—especially human-curated ones—dramatically elevates both lexical fidelity and semantic adequacy. 
Furthermore, multidimensional human preference audits by native experts unveil distinct cross-cultural preferences between Sinographic-tradition and non-Sinographic language groups.
\end{abstract}

\section{Introduction}

Translation is fundamentally an act of cross-cultural negotiation that involves striking a precarious balance between cultural foreignness and target-language accessibility~\cite{aixela20234}. 
As emphasized in recent cross-cultural natural language processing studies~\cite{hershcovich2022challenges,yao2024benchmarking}, the difficulty of faithful cross-lingual transfer largely stems from disparate conceptualizations across societies, wherein language alone often fails to encode implicit regional norms and material traditions. 
While multilingual machine translation (MT) has achieved remarkable progress through massive pre-training~\cite{costa2022no} and the emergence of Large Language Models (LLMs)~\cite{hendy2023good,zhu2024multilingual}, standard architectures remain ill-equipped to negotiate Culturally-Specific Items (CSIs) \cite{villa2025cammt}.
In specialized cultural domains, literal decoding frequently triggers hallucinations, factual inaccuracies, or the outright erasure of indigenous heritage \cite{akinade2023varepsilon,cao2024cultural}. 
This cultural bottleneck is acutely amplified when translating classical cultural heritage, such as historical Yellow River records that document millennia of water conservancy, institutional governance, and hydraulic rituals. 
These texts, composed in terse Classical Chinese, present severe diachronic shifts, historical polysemy, and omitted contextual arguments that challenge the semantic representation of modern MT systems.

\begin{figure*}
    \centering
    \includegraphics[scale=0.39]{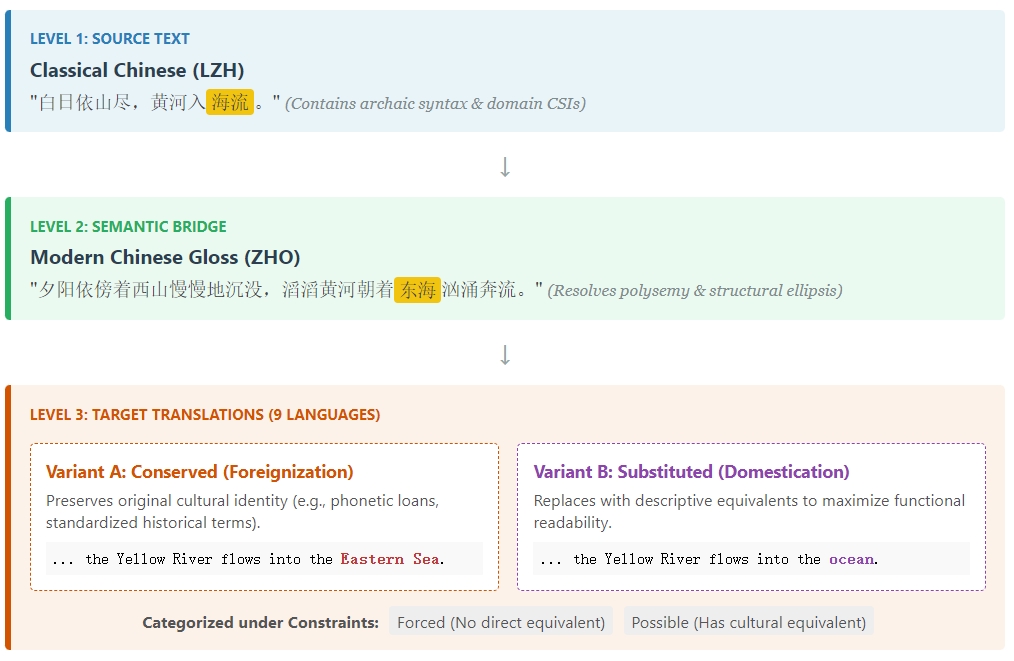}
    \caption{The Tri-Level Architecture and CSI Operationalization in YRC-Bench}
    \label{fig1}
\end{figure*}

Despite nascent efforts to benchmark cultural awareness in MT that ranging from contemporary regional recipes and idioms \cite{cao2024cultural,li2024translate} to multimodal scene-captioning \cite{villa2025cammt,gao2026bookagent}, existing corpora predominantly center on modern standardized tongues and vernacular text genres. 
Conversely, classical language translation research remains largely restricted to bilingual Classical-to-Modern Chinese conversion \cite{liu2024sikugpt,cao2024wenmind}, leaving the direct cross-lingual dissemination of classical heritage into diverse foreign languages fundamentally underexplored. 
In professional human translation workflows, translators rarely map terse classical expressions directly into foreign vernaculars; instead, they routinely rely on intermediate modern paraphrastic glosses to unpack archaic terminology, disambiguate syntactic omissions, and anchor cultural connotations. 
However, whether and how LLMs can systematically capitalize on such intermediate linguistic representations as cognitive and semantic bridges remains an open empirical question.
Furthermore, the handling of specialized classical CSIs has never been rigorously scrutinized under formal translational frameworks that dissect the trade-off between cultural conservation and functional substitution~\citep{chunyuan2024translation,qiufen2025strategies}.

To address these limitations, we introduce \textbf{YRC-Bench}, a comprehensive, human-verified multi-parallel benchmark specifically engineered to evaluate the cross-lingual, culturally aware translation of Classical Chinese heritage. 
As illustrated in Figure~\ref{fig1}, YRC-Bench establishes a tri-level parallel architecture aligning Classical Chinese originals, Modern Chinese explanatory glosses, and expert-verified target translations across nine typologically diverse languages. 
To assess cultural fidelity beyond surface-level metrics, we curate historical CSIs spanning engineering terminology, bureaucratic titles, and ritual customs, formalizing them into paired \textit{Conserved} (retaining foreignized cultural identity) and \textit{Substituted} (domesticated via descriptive or functional equivalents) reference versions under \textit{Forced} and \textit{Possible} translatability conditions. 
Leveraging this framework, we conduct exhaustive bidirectional evaluations across nine representative LLMs under direct, human-bridged, and machine-bridged settings, combining automated metrics with multidimensional human audits.

Our contributions are summarized as follows:
\begin{itemize}
    \item We present YRC-Bench, the first multi-parallel corpus comprising Classical Chinese, Modern Chinese semantic glosses, and nine target foreign languages spanning 23 regions, establishing a rigorous diagnostic suite for historical cultural MT.
    \item We construct a specialized evaluation testbed featuring dual-variant translations (\textit{Conserved} vs. \textit{Substituted}) across hydraulic and institutional CSIs, categorizing cultural equivalence into \textit{Forced} and \textit{Possible} constraints.
    \item Through comprehensive multi-model experiments and native expert evaluations, we demonstrate that intermediate modern glosses significantly improve bidirectional translation quality and CSI recovery. Moreover, our findings reveal systematic preference divergences between Sinographic and non-Sinographic cultural traditions regarding heritage translation strategies.
\end{itemize}

%%%%%%%%%%%%%%%%%%%%%%%%%%%%%%%%%%%%%%%%%%%%%%%%%%%%%%%%%%%%%%%%%%%%%%%%%%%%%%%%%%%%
\section{Related Work}
The cross-lingual translation of historical heritage texts faces dual challenges of diachronic language evolution and cross-cultural knowledge disparities \cite{hershcovich2022challenges}.
While existing research utilizes modern Chinese interpretations as semantic anchors for ancient text translation \cite{liu2024sikugpt,cao2024wenmind,gao2026slvr}, whether these can serve as effective intermediate representations across multiple target languages remains under-investigated. 
Particularly for cultural heritage texts, mapping diachronic semantic transformation to cross-lingual expression within a unified evaluation framework is urgently needed.

Culture-Specific Items (CSIs) are central to this challenge, requiring a trade-off between cultural conservation and substitution \cite{aixela20234, qiufen2025strategies, chunyuan2024translation, cao2024cultural}. Although recent culture-aware machine translation studies evaluate systems using specialized corpora \cite{yao2024benchmarking, villa2025cammt}, translating historical CSIs is further constrained by original historical contexts.
\citet{ma2026musk} proposed multi-scale knowledge learning to jointly capture lexical, sentence-level, and discourse-level cultural semantics.
Inspired by adversarial robustness mechanisms \cite{hu2026research} to improve the reliability of culturally aware translation under ambiguous, misleading, and noisy historical contexts.
Models must make appropriate target-culture expressive choices while accurately restoring native historical semantics.

Furthermore, macroscopic automatic metrics in existing multilingual benchmarks \cite{costa2022no, zhu2024multilingual}—such as lexical overlap \cite{papineni2002bleu, popovic2017chrf++} or pre-trained semantic scores \cite{zhang2019bertscore, rei2020comet}—fail to capture fine-grained cultural concept loss or cross-cultural preference divergences. 
Consequently, this paper introduces YRC-Bench, a novel multilingual historical and cultural translation benchmark. Aligning ancient Chinese, modern Chinese interpretations, and nine foreign languages into a three-tiered parallel corpus, it designs dual CSI variants for conservation and substitution. Through bidirectional evaluation and native-speaker expert assessments, this study systematically investigates the role of explicit semantic interpretations and profiles cross-cultural expression preferences.

%%%%%%%%%%%%%%%%%%%%%%%%%%%%%%%%%%%%%%%%%%%%%%%%%%%%%%%%%%%%%%%%%%%%%%%%%%%%%%%%%%%
\section{The YRC-Bench Dataset}
\label{sec:dataset}

\begin{figure*}
    \centering
    \includegraphics[scale=0.25]{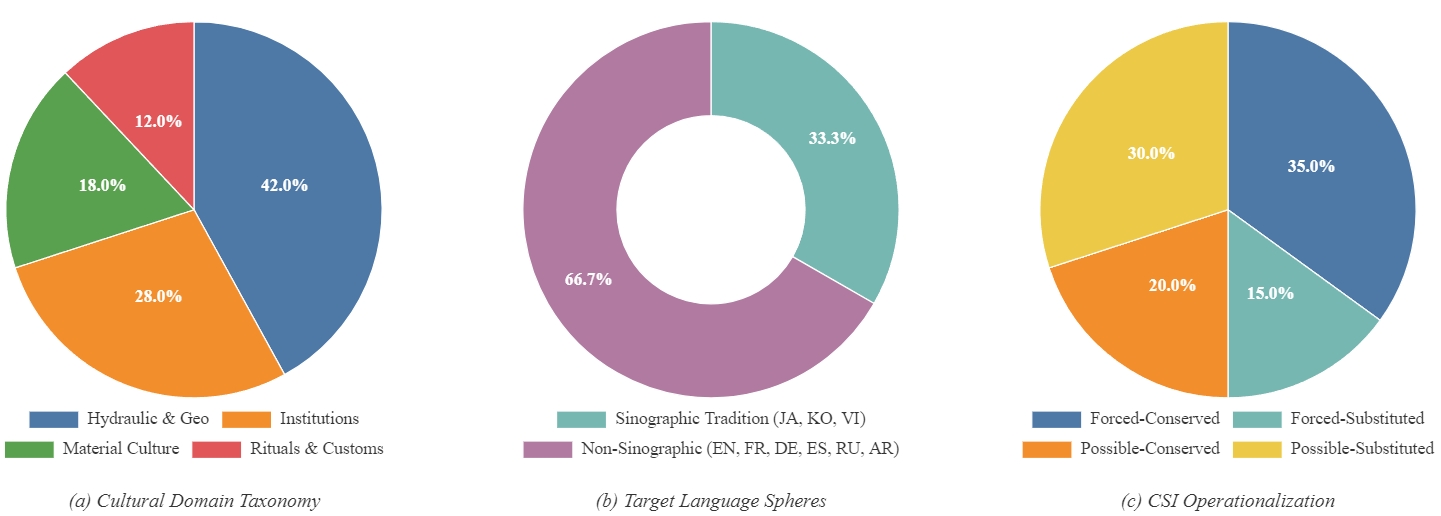}
    \caption{Data Distribution of YRC-Bench Dataset}
    \label{fig2}
\end{figure*}

Constructing a benchmark for classical cultural heritage translation requires reconciling the diachronic divergence between archaic source texts and modern target languages. 
To this end, YRC-Bench is engineered as a multi-parallel, culturally annotated corpus. We detail its curation pipeline, alignment architecture, and CSI operationalization below.

\subsection{Source Data and Cultural Taxonomy}
\label{subsec:source_data}
Data curation commenced with the extraction of authentic historical narratives from foundational Yellow River treatises, notably the \textit{Shuijing Zhu} (Commentary on the Waterways Classic) and \textit{Hefang Yilan} (Overview of River Defense). 
These canonical texts chronicle millennia of hydraulic engineering, socio-political governance, and regional customs.
To capture a comprehensive spectrum of cultural phenomena, we established a targeted taxonomy comprising four primary domains: 
(1) \textit{Hydraulic Engineering \& Geography} (e.g., ancient dyke construction techniques, historical waterways); 
(2) \textit{Institutional Governance} (e.g., bureaucratic titles, taxation systems); 
(3) \textit{Material Culture} (e.g., specialized tools, measuring instruments); and (4) \textit{Rituals \& Customs} (e.g., river-worship ceremonies, local folklore). 
Source passages were meticulously segmented to preserve complete semantic contexts while avoiding excessive text lengths that might confound sentence-level evaluation.

\subsection{Tri-Level Parallel Alignment}
\label{subsec:tri_level}
Directly mapping Classical Chinese (LZH) to foreign languages (FOR) routinely forces models into aggressive hallucination due to systemic syntactic omissions and semantic opacity. 
\citet{zhang2026sainf} intrinsic self-correction mechanisms to enable LLMs to detect cultural semantic errors and selectively refine translation outputs inspiring our work.
We mitigate this by introducing Modern Chinese (ZHO) as an explicit semantic bridge, formulating a \textit{Tri-Level Parallel Alignment} architecture: $\text{LZH} \leftrightarrow \text{ZHO} \leftrightarrow \text{FOR}$. 

The intermediate ZHO layer functions not merely as a literal translation, but as a semantic unpacking mechanism.
It explicitly resolves zero anaphora, explicates archaic polysemy, and provides descriptive context for obscure CSIs.
During construction, professional historical linguists drafted the modern glosses, ensuring absolute fidelity to the classical source. 
Subsequently, bilingual experts translated these glosses into the target languages.
This tri-level structure empowers the benchmark to test models under various bridging conditions, isolating the precise impact of modern semantic mediation on classical text translation.

\subsection{Operationalizing Culturally-Specific Items}
\label{subsec:csi_op}
A distinguishing feature of YRC-Bench is its rigorous operationalization of Culturally-Specific Items.
Borrowing from translation studies on domestication and foreignization, we annotated key CSIs within the corpus and provided two distinct reference variants for the target languages:
\begin{itemize}
    \item \textbf{Conserved Variant (C):} Retains the cultural identity of the source item via transliteration, borrowing, or standard historical nomenclature, prioritizing cultural preservation over immediate accessibility.
    \item \textbf{Substituted Variant (S):} Employs descriptive paraphrasing, functional equivalents, or approximate modern concepts, sacrificing cultural specificity to optimize target-reader comprehension.
\end{itemize}

To further granularize the evaluation, we categorized the translational environment of these CSIs into two constraints based on target-language lexicon availability: 
\textit{Forced} (where the target language lacks any historically proximate equivalent, compelling extreme conservation or heavy description) and \textit{Possible} (where a reasonably adequate cultural equivalent exists). 
This $2 \times 2$ operationalization allows for precise diagnosis of an LLM's cultural translation strategies.

\subsection{Dataset Statistics and Multilingual Coverage}
\label{subsec:statistics}
The finalized YRC-Bench encompasses high-quality parallel triads expanded across nine typologically diverse target languages. 
To investigate the influence of shared historical context on translation behavior, we strategically partitioned the target languages into two distinct cultural spheres: the \textit{Sinographic Tradition} (Japanese, Korean, Vietnamese), which historically shares structural and lexical affinities with Chinese; and the \textit{Non-Sinographic Tradition} (English, French, German, Spanish, Russian, Arabic). 

In total, the dataset yields over 8,500 highly curated evaluation instances (As in Figure \ref{fig2}).
Within these, we identified and annotated 1,240 unique CSIs, yielding over 3,500 targeted CSI evaluation pairs when accounting for the Conserved/Substituted splits. 
The rigorous human validation at every stage ensures that YRC-Bench serves as an uncompromising diagnostic tool for culturally aware machine translation.

%%%%%%%%%%%%%%%%%%%%%%%%%%%%%%%%%%%%%%%%%%%%%%%%%%%%%%%%%%%%%%%%%%%%%%%%%%%%%%%%%%%%
\section{Experimental Setup}
\label{sec:setup}

Our empirical framework isolates the impact of modern semantic mediation on the cross-lingual processing of classical texts. To strictly control for confounding variables, we standardize zero-shot prompts across all translation directions, deferring extensive prompt tuning to future work.

\subsection{Evaluated Models}
\label{subsec:models}
We benchmark nine representative LLMs with proven multilingual proficiency. The selection encompasses proprietary frontier models (e.g., GPT-4o, Claude 3.5 Sonnet) and state-of-the-art open-weight families across various parameter scales (including Qwen2.5, DeepSeek-V2, Seed-X). 
% DeepSeek-V2, Seed-X, and Llama-3). 
To ensure reproducibility, deterministic generation parameters (temperature set to $0.0$) were enforced globally.
An exhaustive list of model versions, context window configurations, and exact prompt templates is provided in Appendix~\ref{app:model_details}.

\subsection{Translation Tasks and Bridging Strategies}
\label{subsec:strategies}
Evaluations span two bidirectional tasks: classical-to-foreign (LZH $\rightarrow$ X) and foreign-to-classical (X $\rightarrow$ LZH). For each direction, we formulate an ablation study featuring three distinct cognitive bridging settings:
\begin{itemize}
    \item \textbf{Direct Translation (DT):} Models map the source text directly to the target language, serving as the baseline for unmediated cross-lingual transfer.
    \item \textbf{Human-Gloss Bridged (HGB):} Models receive the source text accompanied by the gold-standard Modern Chinese gloss from YRC-Bench as auxiliary context prior to generating the target translation.
    \item \textbf{Machine-Gloss Bridged (MGB):} Models operate under a zero-shot Chain-of-Thought (CoT) paradigm, autonomously generating an intermediate Modern Chinese paraphrase before finalizing the translation. This setting probes the models' intrinsic capability to construct their own semantic hubs.
\end{itemize}

\subsection{Evaluation Metrics}
\label{subsec:metrics}
We assess translation quality through a dual-lens approach, capturing both holistic semantic alignment and localized cultural fidelity. 

For overall text quality, we report BLEU~\citep{papineni2002bleu} and chrF++~\citep{popovic2017chrf++} for surface-level lexical overlap, alongside COMET~\citep{rei2020comet} and BERTScore~\citep{zhang2019bertscore} for neural semantic equivalence. 

To quantify the treatment of cultural nuances, we introduce two specialized metrics tailored to the dual-variant CSIs defined in Section~\ref{sec:dataset}:
\begin{enumerate}
    \item \textbf{Cultural Retention Rate (CRR):} Evaluated against the \textit{Conserved} reference subset, this metric measures the frequency at which models successfully preserve foreignized cultural terminology without reverting to generic descriptions.
    \item \textbf{Cultural Recovery Rate (CoRR):} Evaluated against the \textit{Substituted} reference subset, this measures whether models can recover the precise cultural referent when prompted with a domesticated or descriptive input (e.g., reverting a functional description in English back to its specific classical title in Chinese).
\end{enumerate}

%%%%%%%%%%%%%%%%%%%%%%%%%%%%%%%%%%%%%%%%%%%%%%%%%%%%%%%%%%%%%%%%%%%%%%%%%%%%%%%%%

\begin{table}[t]
\centering
\scriptsize
\begin{tabular}{lcccc}
\toprule
\multirow{2}{*}{\textbf{Model}} & \multicolumn{2}{c}{\textbf{LZH $\rightarrow$ X}} & \multicolumn{2}{c}{\textbf{X $\rightarrow$ LZH}} \\
\cmidrule(lr){2-3} \cmidrule(lr){4-5}
 & \textbf{chrF++} & \textbf{COMET} & \textbf{chrF++} & \textbf{COMET} \\
\midrule
GPT-4o & 43.12 & 0.785 & 35.67 & 0.812 \\
Claude 3.5 & \textbf{44.05} & \textbf{0.792} & \textbf{36.81} & \textbf{0.825} \\
Qwen2.5-72B & 41.88 & 0.764 & 34.22 & 0.798 \\
DeepSeek-V2 & 40.52 & 0.751 & 33.15 & 0.784 \\
\bottomrule
\end{tabular}
\caption{Baseline Direct Translation (DT) performance averaged across all target languages. Best results in bold.}
\label{tab:main_results}
\end{table}

\section{Quantitative Results and Analysis}
\label{sec:results}

We report the evaluation outcomes across bidirectional translation tracks, assessing the efficacy of semantic bridging and the resolution of culturally-specific items (CSIs). 
The analysis uncovers the mechanical limitations of direct classical translation and elucidates why modern semantic hubs mitigate these bottlenecks.

\subsection{Overall Bidirectional Translation Performance}
\label{subsec:overall_perf}

Table~\ref{tab:main_results} details the baseline performance of Direct Translation (DT) across four representative frontiers and open-weight models.
Unmediated translation of Classical Chinese (LZH) imposes severe cognitive loads on models, reflected in uniformly suppressed COMET and chrF++ scores. 
Specifically, the LZH $\rightarrow$ X direction exhibits a pronounced performance floor compared to modern language benchmarks. 

This systemic degradation stems from the syntactic density of LZH where extensive zero anaphora and omission of logical connectives force models into aggressive guesswork, frequently triggering localized hallucinations. 
Conversely, the X $\rightarrow$ LZH trajectory yields marginally better neural semantic overlap (COMET), likely because generating LZH relies heavily on stylistic replication, a task at which instruction-tuned models excel, even if the underlying semantic mapping remains fragile.

\begin{figure}[t]
    \centering
    \includegraphics[scale=0.29]{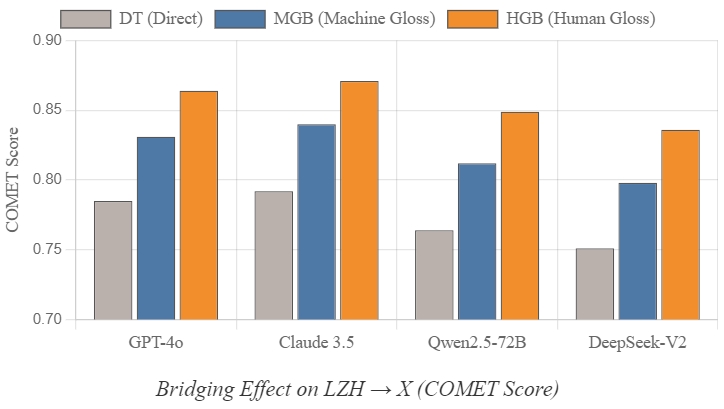}
    \caption{Ablation of Semantic Bridging Efficacy Across Models}
    \label{fig3a}
\end{figure}

\subsection{Main Results}
\label{subsec:main_result}
Table~\ref{tab:translation_results} demonstrates that human-provided Modern Chinese glosses generally enhance translation performance, with more pronounced gains observed in the X$\rightarrow$LZH direction. However, Qwen3-8B exhibits performance degradation under HGB for LZH$\rightarrow$X, indicating that the efficacy of these glosses is contingent upon both the model and the translation direction. Conversely, MGB yields marginal and inconsistent improvements; in X$\rightarrow$LZH, its lexical gains for most models coincide with a decline in BERTScore across all evaluated models. These findings suggest that while reference glosses serve as valuable auxiliary context, prompting models to generate intermediate paraphrases fails to reliably replicate these benefits within the tested zero-shot framework.

\begin{table*}[t]
\centering
\small
\setlength{\tabcolsep}{5pt}
\renewcommand{\arraystretch}{1.2}

\resizebox{0.9\textwidth}{!}{
\begin{tabular}{llcccccccc}
\toprule

\multicolumn{2}{c}{\multirow{2}{*}{Model}}
& \multicolumn{4}{c}{LZH-X}
& \multicolumn{4}{c}{X-LZH} \\

\cmidrule(lr){3-6}
\cmidrule(lr){7-10}

\multicolumn{2}{c}{}
& BLEU & ChrF++ & BERTScore & COMET
& BLEU & ChrF++ & BERTScore & COMET \\

\midrule

% Qwen3.8-27B
\multirow{3}{*}{Qwen3.8-27B}
& DT  & 26.8670 & 41.7430 & 0.8374 & 0.778124 & 20.5774 & 16.8792 & 0.8443 & 0.788221 \\
& HGB & \textbf{30.2943} & \textbf{45.6224} & \textbf{0.8533} & \textbf{0.804146} & \textbf{37.9433} & \textbf{28.7248} & \textbf{0.8861} & \textbf{0.824925} \\
& MGB & 27.3879 & 42.9405 & 0.8416 & 0.791304 & 21.3519 & 17.5757 & 0.8393 & 0.786278 \\

\midrule

% Gemma3-27B
\multirow{3}{*}{Gemma3-27B}
& DT  & 21.0019 & 36.0892 & 0.8128 & 0.736109 & 11.6991 & 11.3138 & 0.8243 & 0.768620 \\
& HGB & \textbf{25.7769} & \textbf{41.0833} & \textbf{0.8344} & \textbf{0.775935} & \textbf{18.3080} & \textbf{16.1042} & \textbf{0.8539} & \textbf{0.800608} \\
& MGB & 21.6502 & 36.6112 & 0.8157 & 0.750314 & 10.6882 & 11.1445 & 0.8174 & 0.765360 \\

\midrule

% Gemini3.1
\multirow{3}{*}{Gemini3.1}
& DT  & 31.7368 & 45.6854 & 0.8511 & 0.799644 & 19.2446 & 16.4392 & 0.8542 & 0.801992 \\
& HGB & \textbf{32.8266} & \textbf{47.7065} & \textbf{0.8589} & \textbf{0.821210} & \textbf{35.8342} & \textbf{28.6618} & \textbf{0.8896} & \textbf{0.846840} \\
& MGB & 31.1692 & 46.0854 & 0.8517 & 0.810298 & 21.8842 & 18.5324 & 0.8491 & 0.801984 \\

\midrule

% Qwen3-8B
\multirow{3}{*}{Qwen3-8B}
& DT  & \textbf{16.4975} & \textbf{32.4854} & \textbf{0.8008} & \textbf{0.711960} & 12.2196 & 11.5365 & 0.8101 & 0.759374 \\
& HGB & 12.2149 & 24.3329 & 0.7734 & 0.645202 & \textbf{29.0430} & \textbf{23.8852} & \textbf{0.8632} & \textbf{0.814734} \\
& MGB & 11.3514 & 23.2103 & 0.7673 & 0.631280 & 12.9073 & 12.4506 & 0.8031 & 0.765115 \\

\bottomrule
\end{tabular}
}

\caption{Translation performance comparison of four models under three experimental settings (DT, HGB, and MGB) in both LZH-X and X-LZH translation directions. Best results in bold.}

\label{tab:translation_results}

\end{table*}

\subsection{The Bridging Effect of Modern Chinese}
\label{subsec:bridging_effect}

Introducing ZHO as a mediating layer dramatically alters the translation dynamics. 
As depicted in Figure \ref{fig3a}, the HGB setting yields absolute COMET improvements of up to $8.5\%$ over the DT baseline across all models. 
This surge confirms the theoretical premise: ZHO functions as a highly effective semantic hub, unpacking archaic opacity into standardized syntactic structures that align perfectly with the models' dense pre-training distributions.

The MGB paradigm, operating via zero-shot autonomous paraphrasing, also outperforms the DT baseline but suffers a noticeable margin of degradation compared to HGB. This discrepancy exposes the vulnerability of autonomous decoding: error propagation. When a model misinterprets the classical source during the initial LZH $\rightarrow$ ZHO generation phase, the semantic distortion cascades irreparably into the final foreign translation. Consequently, while LLMs possess the structural capacity for intermediate reasoning, human-curated glosses remain indispensable for anchoring absolute historical fidelity.

\begin{figure}[t]
    \centering
    \includegraphics[scale=0.29]{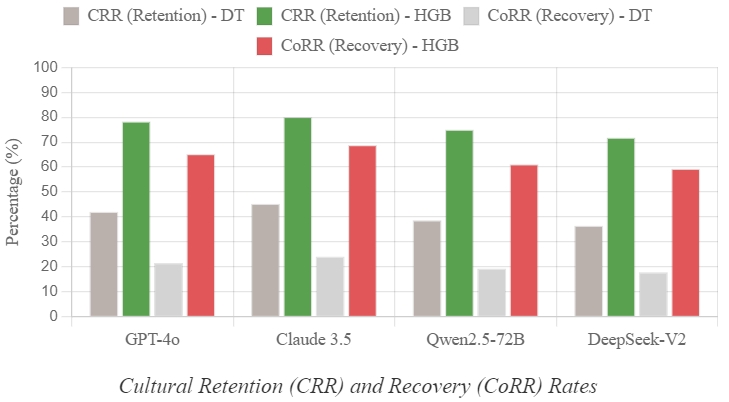}
    \caption{Ablation of CSI Resolution Efficacy Across Models}
    \label{fig3b}
\end{figure}

\subsection{Performance on Culturally-Specific Items}
\label{subsec:csi_perf}

The treatment of historical terminology provides the most granular lens into cultural awareness.
Figure \ref{fig3b} illustrates the CRR and CoRR under varying bridging constraints. 
Under the DT setup, models struggle to retain foreignized CSIs (low CRR), typically defaulting to over-domestication or semantic dropping. 
When explicitly bridged by HGB, CRR spikes significantly. 
The modern gloss acts as an explicit semantic boundary, instructing the model exactly which entity must be preserved versus which context can be freely adapted. 

More critically, in the X $\rightarrow$ LZH direction, recovering a specific classical title from a domesticated foreign description (CoRR) proves near-impossible for direct translation due to the severe information asymmetry. 
The ZHO bridge narrows this gap by mapping the foreign descriptive phrase to a modern Chinese equivalent first, from which the model can probabilistically infer the correct classical nomenclature.
Ultimately, the results demonstrate that cultural translation is less about brute-force multilingual mapping and more about step-wise semantic triangulation.

%%%%%%%%%%%%%%%%%%%%%%%%%%%%%%%%%%%%%%%%%%%%%%%%%%%%%%%%%%%%%%%%%%%%%%%%%%%%%%%%%%%
\section{Multidimensional Human Audit and Preferences}
\label{sec:human_audit}

Automated metrics intrinsically lack the granularity to capture subtle cultural resonance, stylistic dissonance, and contextual adequacy. To rigorously validate our quantitative findings and probe the sociolinguistic behaviors of models, we conducted a comprehensive human audit engaging bilingual native experts across the nine target languages.

\subsection{Multidimensional Translation Quality Audit}
\label{subsec:multi_audit}
We established a five-dimensional human evaluation rubric comprising \textit{Accuracy} (lexical and factual correctness), \textit{Completeness} (absence of unwarranted omissions), \textit{Culture} (faithful preservation of historical connotations), \textit{Fluency} (target-language naturalness), and \textit{Style} (appropriateness of historical register). Evaluators conducted blind scoring on a subset of 1,000 diverse instances.

% \begin{table}[t]
% \centering
% \begin{tabular}{lccccc}
% \toprule
% \textbf{Setting} & \textbf{Acc.} & \textbf{Comp.} & \textbf{Cult.} & \textbf{Flu.} & \textbf{Style} \\
% \midrule
% \multicolumn{6}{c}{\textit{Direction: LZH $\rightarrow$ X}} \\
% \midrule
% DT & 2.45 & 2.88 & 1.95 & 3.42 & 2.15 \\
% MGB & 3.52 & 3.81 & 3.10 & 3.75 & 3.22 \\
% HGB & \textbf{4.35} & \textbf{4.62} & \textbf{4.45} & \textbf{4.30} & \textbf{4.18} \\
% \midrule
% \multicolumn{6}{c}{\textit{Direction: X $\rightarrow$ LZH}} \\
% \midrule
% DT & 1.85 & 2.55 & 1.62 & 2.85 & 1.55 \\
% MGB & 3.15 & 3.42 & 2.88 & 3.55 & 2.75 \\
% HGB & \textbf{4.12} & \textbf{4.35} & \textbf{4.20} & \textbf{4.15} & \textbf{4.05} \\
% \bottomrule
% \end{tabular}
% \caption{Average multidimensional human audit scores (1-5 scale) across all target languages. Human-Gloss Bridged (HGB) exhibits overwhelming superiority, particularly rescuing the \textit{Culture} and \textit{Style} dimensions which suffer fatal degradation under Direct Translation (DT).}
% \label{tab:human_audit}
% \end{table}

\begin{table}[t]
\centering
\small
\setlength{\tabcolsep}{3pt}
\renewcommand{\arraystretch}{1.0}

\begin{tabular}{lccccc}
\toprule
\textbf{Setting} & \textbf{Acc.} & \textbf{Comp.} & \textbf{Cult.} & \textbf{Flu.} & \textbf{Style} \\
\midrule
\multicolumn{6}{c}{\textit{Direction: LZH $\rightarrow$ X}} \\
\midrule
DT  & 2.45 & 2.88 & 1.95 & 3.42 & 2.15 \\
MGB & 3.52 & 3.81 & 3.10 & 3.75 & 3.22 \\
HGB & \textbf{4.35} & \textbf{4.62} & \textbf{4.45} & \textbf{4.30} & \textbf{4.18} \\
\midrule
\multicolumn{6}{c}{\textit{Direction: X $\rightarrow$ LZH}} \\
\midrule
DT  & 1.85 & 2.55 & 1.62 & 2.85 & 1.55 \\
MGB & 3.15 & 3.42 & 2.88 & 3.55 & 2.75 \\
HGB & \textbf{4.12} & \textbf{4.35} & \textbf{4.20} & \textbf{4.15} & \textbf{4.05} \\
\bottomrule
\end{tabular}

\caption{Average multidimensional human audit scores (1-5 scale) across all target languages. Human-Gloss Bridged (HGB) exhibits overwhelming superiority, particularly rescuing the \textit{Culture} and \textit{Style} dimensions which suffer fatal degradation under Direct Translation (DT). Best results in bold.}
\label{tab:human_audit}
\end{table}

% \begin{table*}[ht]
% \centering
% \begin{tabular}{lcccc}
% \toprule
% \multirow{2}{*}{\textbf{Dimension}} & \multicolumn{2}{c}{\textbf{X $\rightarrow$ LZH}} & \multicolumn{2}{c}{\textbf{LZH $\rightarrow$ X}} \\
% \cmidrule(lr){2-3} \cmidrule(lr){4-5}
%  & \textbf{Count ($T+M$/$T$)} & \textbf{Win\%} & \textbf{Count ($T+M$/$T$)} & \textbf{Win\%} \\
% \midrule
% Accuracy     & 1012 / 428 & 69.6\% & 958 / 475 & 66.4\% \\
% Completeness & 830 / 345  & 66.3\% & 815 / 352 & 65.7\% \\
% Culture      & 610 / 126  & 68.3\% & 414 / 201 & 58.2\% \\
% Fluency      & 865 / 444  & 64.1\% & 869 / 483 & 63.1\% \\
% Style        & 924 / 416  & 67.1\% & 762 / 361 & 63.6\% \\
% \bottomrule
% \end{tabular}
% \caption{Multidimensional pairwise preference breakdown between bridged ($T+M$) and unbridged ($T$) outputs. Win\% denotes the preference share for $T+M$.}
% \label{tab:dimension_preference}
% \end{table*}

\begin{table*}[ht]
\centering
\small
\setlength{\tabcolsep}{3pt}
\renewcommand{\arraystretch}{1.0}

\begin{tabular}{lcccc}
\toprule
\multirow{2}{*}{\textbf{Dimension}}
& \multicolumn{2}{c}{\textbf{X $\rightarrow$ LZH}}
& \multicolumn{2}{c}{\textbf{LZH $\rightarrow$ X}} \\
\cmidrule(lr){2-3}
\cmidrule(lr){4-5}

& \textbf{Count ($T+M$/$T$)} & \textbf{Win\%}
& \textbf{Count ($T+M$/$T$)} & \textbf{Win\%} \\
\midrule

Accuracy     & 1012 / 428 & 69.6\% & 958 / 475 & 66.4\% \\
Completeness & 830 / 345  & 66.3\% & 815 / 352 & 65.7\% \\
Culture      & 610 / 126  & 68.3\% & 414 / 201 & 58.2\% \\
Fluency      & 865 / 444  & 64.1\% & 869 / 483 & 63.1\% \\
Style        & 924 / 416  & 67.1\% & 762 / 361 & 63.6\% \\

\bottomrule
\end{tabular}

\caption{Multidimensional pairwise preference breakdown
between bridged ($T+M$) and unbridged ($T$) outputs.
Win\% denotes the preference share for $T+M$.}
\label{tab:dimension_preference}
\end{table*}

\begin{figure}[t]
    \centering
    \includegraphics[scale=0.16]{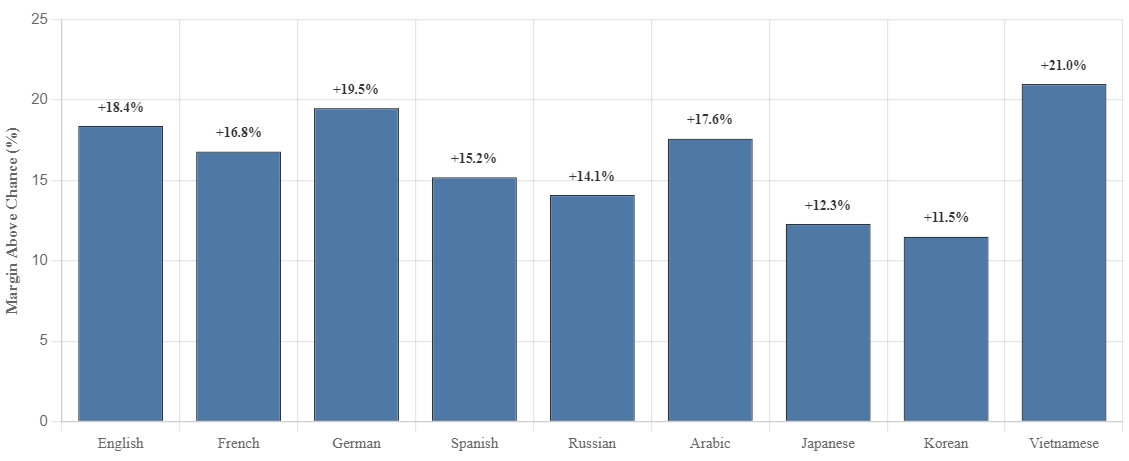}
    \caption{Net Human Preference Win-Rate of Bridged (T+M) over Direct (T) across Target Languages}
    \label{fig4}
\end{figure}

As illustrated in Table~\ref{tab:human_audit}, Direct Translation (DT) collapses completely under the \textit{Culture} and \textit{Style} dimensions, corroborating the hypothesized difficulty of unmediated diachronic mapping. The models frequently generate anachronistic phrasing or erase institutional nuances. HGB effectively eliminates these deficits. The experts noted that modern glosses provide an explicit semantic scaffold, enabling models to concentrate their inferential capacity on refining target-language stylistic elegance rather than struggling to decode the archaic source.

\subsection{Native Expert Preferences on Bridging}
\label{subsec:preferences}
To rigorously validate the translational utility of intermediate glosses from an end-user perspective, we conducted pairwise preference evaluations comparing direct translation ($T$) with human-gloss-bridged translation ($T+M$). 
Table~\ref{tab:dimension_preference} presents the dimensional breakdown of expert choices across both translation directions, while Figure~\ref{fig4} visualizes the net win-rate above chance across all target languages.

% \begin{table*}[t]
% \centering
% \begin{tabular}{lcccc}
% \toprule
% \multirow{2}{*}{\textbf{Tradition}} & \multicolumn{2}{c}{\textbf{Forced Condition}} & \multicolumn{2}{c}{\textbf{Possible Condition}} \\
% \cmidrule(lr){2-3} \cmidrule(lr){4-5}
%  & \textbf{Conserved (\%)} & \textbf{Substituted (\%)} & \textbf{Conserved (\%)} & \textbf{Substituted (\%)} \\
% \midrule
% Sinographic     & $57.83 \pm 21.19$ & $42.17 \pm 21.19$ & $66.38 \pm 7.55$ & $33.62 \pm 7.55$ \\
% Non-Sinographic & $78.56 \pm 18.80$ & $21.44 \pm 18.80$ & $20.76 \pm 7.32$ & $79.24 \pm 7.32$ \\
% \bottomrule
% \end{tabular}
% \caption{Expert translation strategy choices (\%) under Forced and Possible CSI conditions across Sinographic (JA, KO, VI) and Non-Sinographic traditions.}
% \label{tab:csi_tradition}
% \end{table*}

\begin{table*}[t]
\centering
\small
\setlength{\tabcolsep}{3pt}
\renewcommand{\arraystretch}{1.0}
\begin{tabular}{lcccc}
\toprule
\multirow{2}{*}{\textbf{Tradition}}
& \multicolumn{2}{c}{\textbf{Forced Condition}}
& \multicolumn{2}{c}{\textbf{Possible Condition}} \\
\cmidrule(lr){2-3}
\cmidrule(lr){4-5}
& \textbf{Conserved (\%)} & \textbf{Substituted (\%)}
& \textbf{Conserved (\%)} & \textbf{Substituted (\%)} \\
\midrule

Sinographic
& $57.83 \pm 21.19$ & $42.17 \pm 21.19$
& $66.38 \pm 7.55$ & $33.62 \pm 7.55$ \\

Non-Sinographic
& $78.56 \pm 18.80$ & $21.44 \pm 18.80$
& $20.76 \pm 7.32$ & $79.24 \pm 7.32$ \\

\bottomrule
\end{tabular}

\caption{Expert translation strategy choices (\%)
under Forced and Possible CSI conditions across
Sinographic (JA, KO, VI) and Non-Sinographic traditions.}
\label{tab:csi_tradition}
\end{table*}

The empirical tallies decisively corroborate the bridging hypothesis across all evaluative axes. 
In the reverse direction (X $\rightarrow$ LZH), where models are prone to archaic syntax breakdown, semantic bridging achieves an overwhelming 69.6\% win-rate in \textit{Accuracy} ($1012/428$ selections) and 68.3\% in \textit{Culture}. 
The modern gloss supplies missing contextual arguments, preventing spurious lexical substitutions for historical entities. 
In the LZH $\rightarrow$ X direction, \textit{Accuracy} (66.4\%) and \textit{Completeness} (65.7\%) similarly dominate, demonstrating that intermediate semantic explicitation fundamentally curbs information truncation. 
As reflected in Figure~\ref{fig4}, positive margins are observed across all nine target languages, with distant European languages and Vietnamese exhibiting the most pronounced gains ($>15\%$ above chance).

\subsection{Cross-Cultural Discrepancies: Sinographic vs. Non-Sinographic}
\label{subsec:cultural_discrepancy}

\begin{figure}[ht]
    \centering
    \includegraphics[scale=0.16]{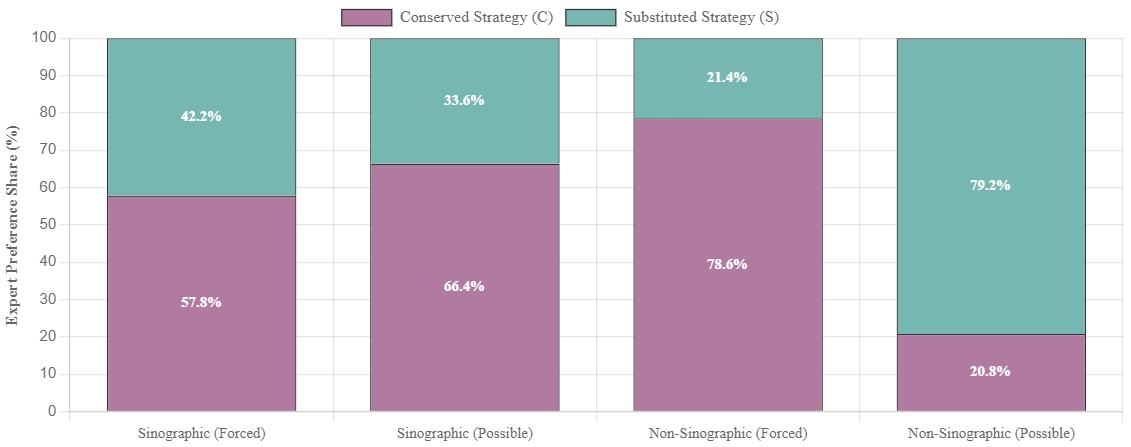}
    \caption{Translation Strategy Preference Share (\%) under Forced and Possible Conditions}
    \label{fig5}
\end{figure}

Beyond bridging utility, our framework probes how divergent linguistic heritage shapes translation strategies for Culturally-Specific Items (CSIs). Table~\ref{tab:csi_tradition} and Figure~\ref{fig5} report native expert choices between \textit{Conserved} (C) and \textit{Substituted} (S) strategies under \textit{Forced} versus \textit{Possible} translatability conditions, grouped by cultural spheres.

The empirical distribution exposes profound cross-cultural divergences. Under the \textit{Possible} condition where target equivalents exist evaluators from the Sinographic tradition decidedly favor \textit{Conserved} variants ($66.38\% \pm 7.55\%$).
In contrast, Non-Sinographic experts overwhelmingly reject conservation, favoring \textit{Substituted} equivalents ($79.24\% \pm 7.32\%$). 
This striking dichotomy reflects the cognitive barrier of cultural reception: shared morphosyntactic and etymological roots (e.g., Hanja, Kanji, and Hán-Nôm) allow Sinographic readers to parse foreignized classical terms intuitively, whereas Non-Sinographic readers experience severe cognitive alienation unless familiar domestic equivalents are deployed.

Conversely, under the \textit{Forced} condition, Non-Sinographic experts pivot sharply toward \textit{Conserved} strategies ($78.56\% \pm 18.80\%$), whereas Sinographic experts exhibit a more balanced distribution ($57.83\%$ vs. $42.17\%$). Qualitative audits explain this inversion: when no authentic cultural equivalent exists, ad-hoc descriptive substitutions in Western languages frequently introduce anachronisms or gross factual errors. Non-Sinographic experts consequently prefer foreignized borrowing to protect historical rigor, while Sinographic annotators remain comfortable with flexible calques. 
These findings substantiate that MT models cannot adopt a uniform CSI strategy; decoding prompts must be culturally parameterized according to target orthographic traditions.
More importantly, conservation preferences among non-Sinographic experts shift by 57.80 percentage points between Forced and Possible conditions, compared with only 8.55 points among Sinographic experts. 
This contrast suggests that cultural translation strategies depend on the interaction between linguistic tradition and the availability of suitable cultural equivalents, rather than a fixed preference for either strategy.

%%%%%%%%%%%%%%%%%%%%%%%%%%%%%%%%%%%%%%%%%%%%%%%%%%%%%%%%%%%%%%%%%%%%%%%%%%%%%%%%%%%%
\section{Conclusion}

In this paper, we introduced \textbf{YRC-Bench}, a multi-parallel benchmark dedicated to diagnosing culturally aware machine translation for Classical Chinese historical heritage across nine typologically diverse target languages. 
By operationalizing a tri-level alignment framework coupled with dual-variant CSI reference sets, our empirical evaluations demonstrate that intermediate modern glosses act as an indispensable cognitive bridge, dramatically resolving archaic syntactic parataxis, suppressing model hallucination, and bolstering both localized cultural term recovery and global holistic adequacy in bidirectional translation.
Furthermore, our multidimensional native expert audits reveal profound cross-cultural fault lines between Sinographic and Non-Sinographic traditions, establishing that optimal heritage translation cannot rely on a monolithic decoding prompt but strictly mandates culturally parameterized strategies tailored to the target language's historical proximity.

\clearpage
\section*{Limitations}
As establishing expert-verified tri-level alignments and granular CSI categorizations requires scarce historical linguistics and philological expertise, the dataset's scale is inherently constrained compared to massive web-crawled modern corpora, primarily positioning YRC-Bench as an evaluation benchmark rather than a large-scale pre-training corpus. 
Then, while the selected Yellow River treatises span rich hydraulic, administrative, and ritual terminology, our taxonomy does not encompass other classical literary genres, such as metered poetry or colloquial historical fiction, where figurative metaphors and rhyme schemes introduce distinct translational bottlenecks

% Bibliography entries for the entire Anthology, followed by custom entries
%\bibliography{custom,anthology-overleaf-1,anthology-overleaf-2}

% Custom bibliography entries only
\bibliography{custom}

\appendix

\section{Appendix}
\subsection{Prompt and Task instruction}
\label{sec:appendix}
In ours experiments, we use two types of prompts for translation tasks: Direct Translation (DT) and Human-Gloss Bridged (HGB/MGB). The prompts are defined as follows:
\begin{lstlisting}
PROMPT_DT = '''Translate the following sentence from {source} to {target}. Provide ONLY the translated text, with no additional information, explanation, or context.
"{sentence}"
'''
\end{lstlisting}

\begin{lstlisting}
PROMPT_HGB = '''Translate the following sentence from {source} to {target} using the provided Modern Chinese sentence as additional context. Provide ONLY the translated text, with no additional information, explanation, or context.
Modern Chinese context: "{modern_chinese}"
"{sentence}"
'''
\end{lstlisting}

\includegraphics[width=0.95\linewidth]{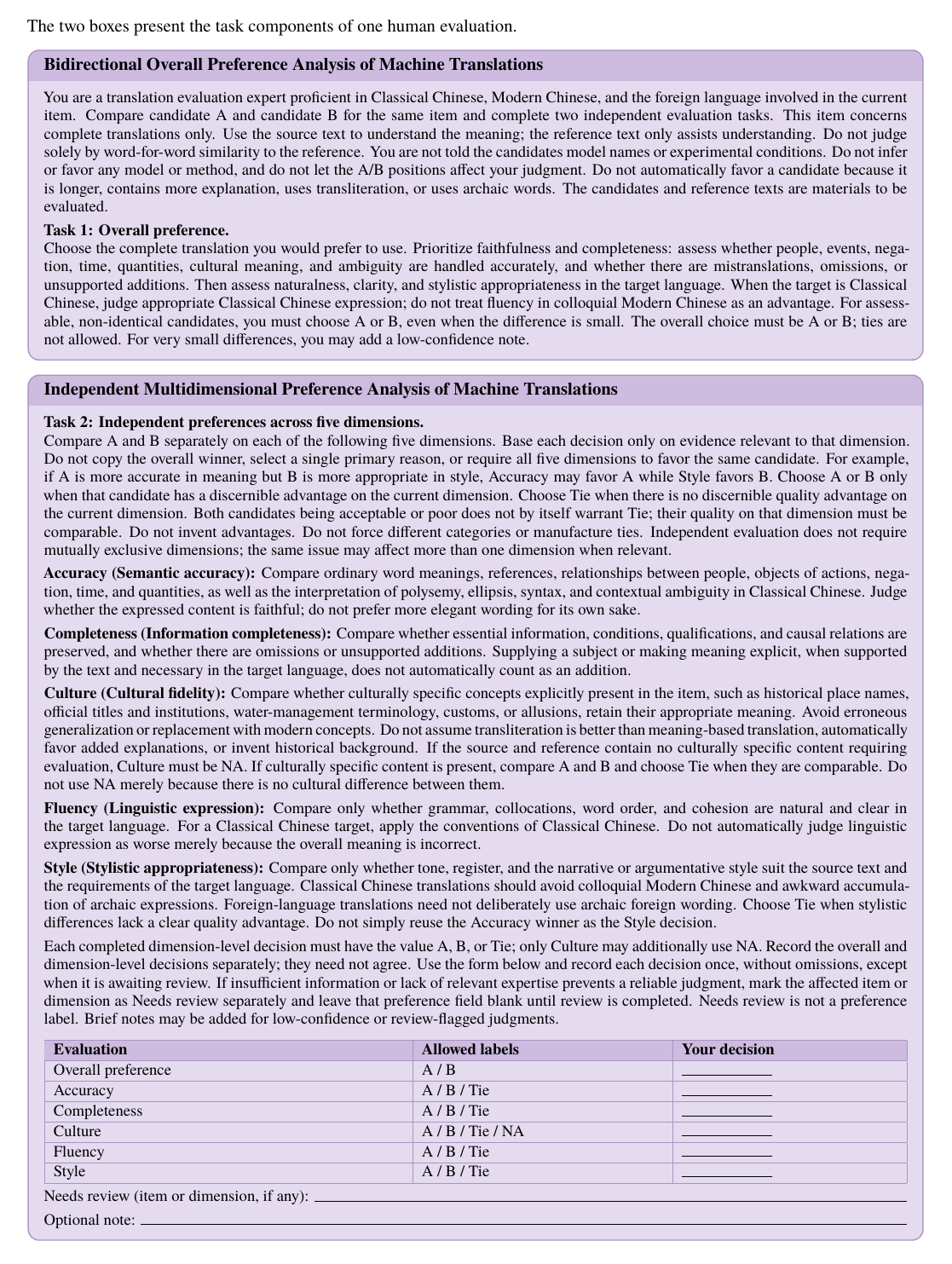}

\includegraphics[width=0.95\linewidth]{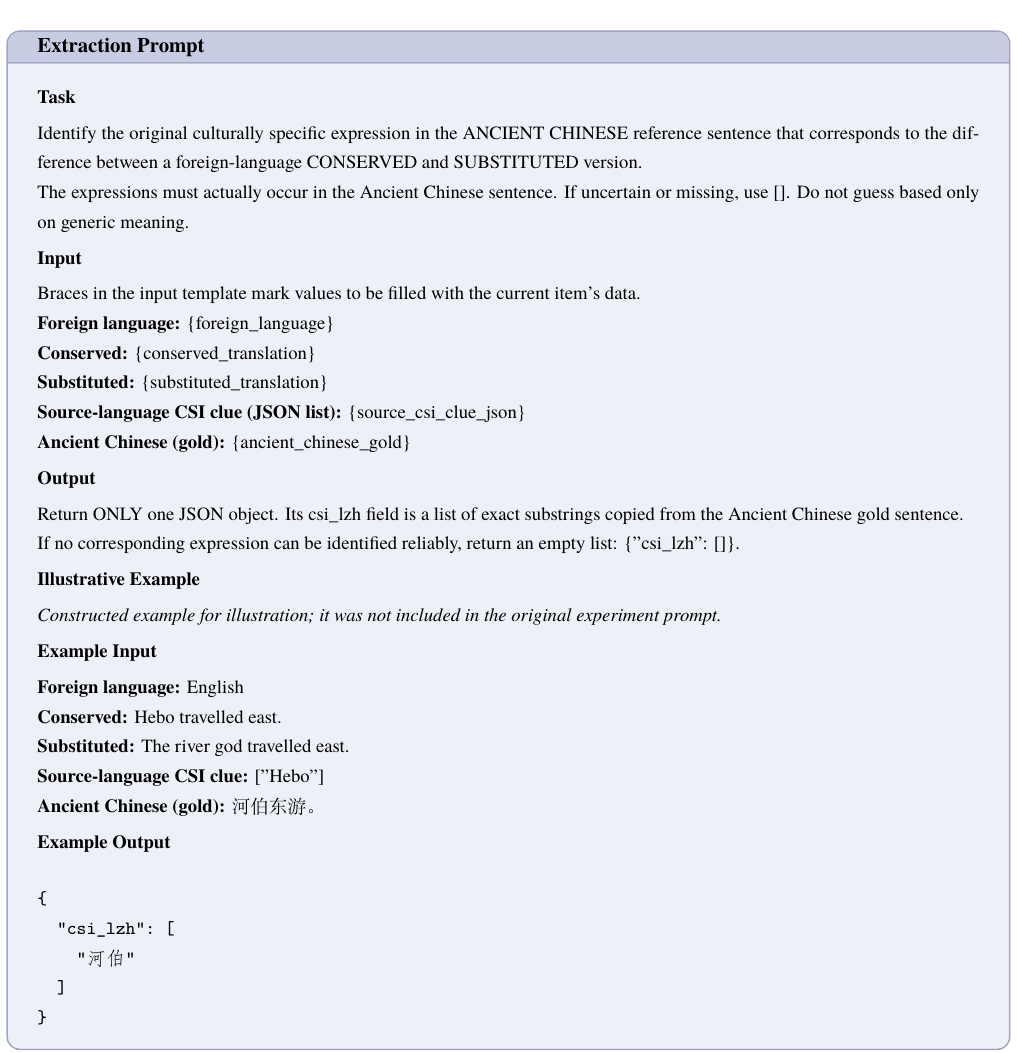}
    
\includegraphics[width=0.95\linewidth]{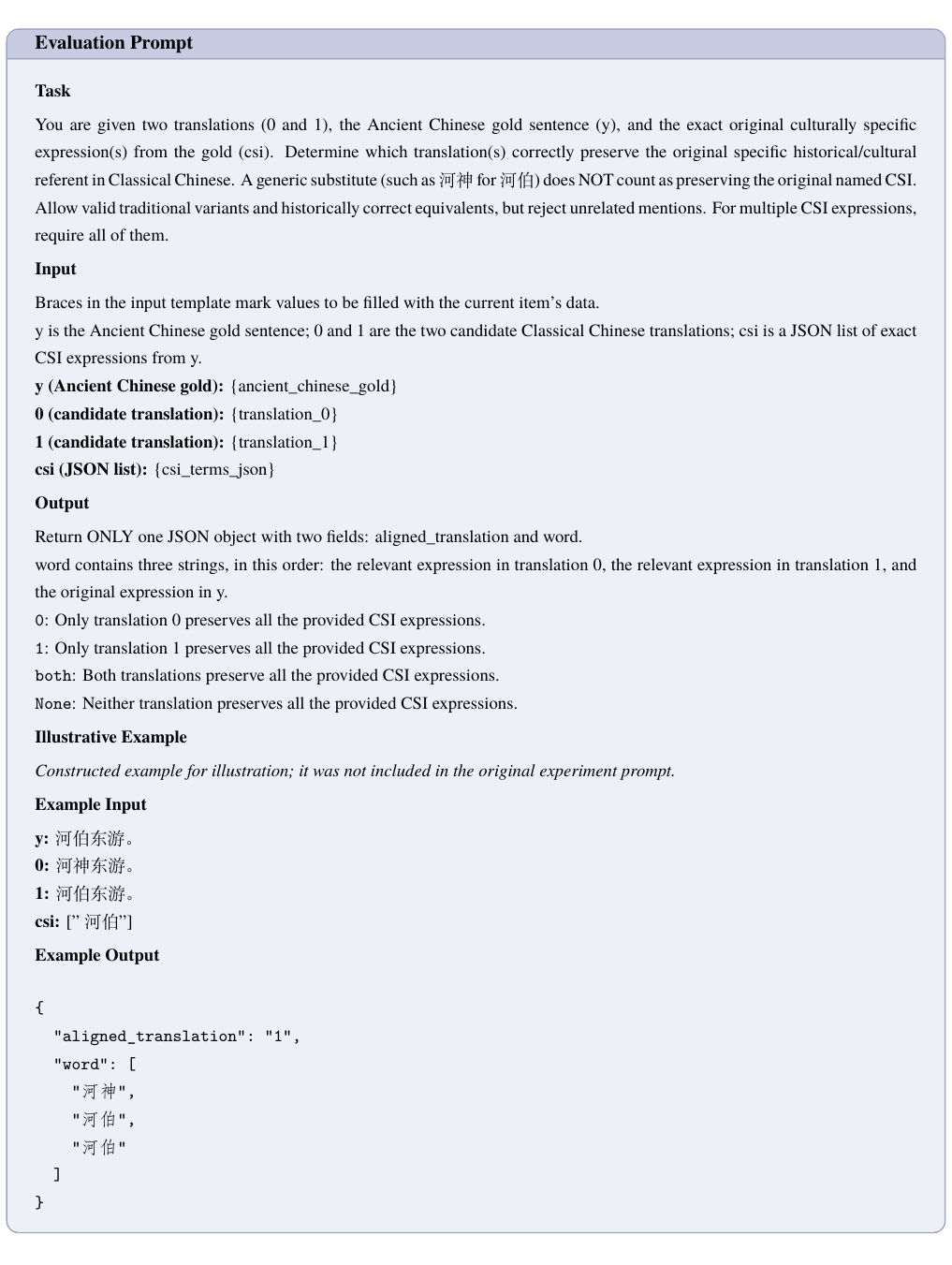}

\subsection{Heatmap Analysis}
\label{Heatmap}
% Required packages in the preamble:
% \usepackage{booktabs}
% \usepackage{multirow}
% \usepackage{graphicx}

% \begin{table*}[t]
% \centering
% \small
% \setlength{\tabcolsep}{5pt}
% \renewcommand{\arraystretch}{1.2}

% \resizebox{\textwidth}{!}{
% \begin{tabular}{lcccccccc}
% \toprule

% \multicolumn{1}{c}{\multirow{2}{*}{Model(DI)}}
% & \multicolumn{4}{c}{LZH-X}
% & \multicolumn{4}{c}{X-LZH} \\

% \cmidrule(lr){2-5}
% \cmidrule(lr){6-9}

% & BLEU & ChrF++ & BERTScore & COMET
% & BLEU & ChrF++ & BERTScore & COMET \\

% \midrule

% Qwen3-8B
% & 16.4975 & 32.4854 & 0.8008 & 0.711960
% & 12.2196 & 11.5365 & 0.8101 & 0.759374 \\

% Qwen3.8-27B
% & 26.8670 & 41.7430 & 0.8374 & 0.778124
% & 20.5774 & 16.8792 & 0.8443 & 0.788221 \\

% Seed-X-PPO
% & 16.0290 & 41.8972 & 0.8083 & 0.728496
% & 6.8269 & 8.9710 & 0.7342 & 0.750249 \\

% Seed-X-Instruct
% & 14.6113 & 31.3028 & 0.7834 & 0.730740
% & 5.9654 & 9.5262 & 0.7247 & 0.720437 \\

% Qwen3.7
% & 26.2831 & 41.1017 & 0.8373 & 0.777979
% & 23.9789 & 19.5319 & 0.8444 & 0.789696 \\

% Deepseek-flash
% & 31.5811 & 45.2177 & 0.8477 & 0.780763
% & 25.9394 & 20.7149 & 0.8735 & 0.811545 \\

% GPT5.5
% & 36.6704 & 49.2389 & 0.8642 & 0.804275
% & 21.7916 & 17.7811 & 0.8659 & 0.807057 \\

% Gemini3.1
% & 31.7368 & 45.6854 & 0.8511 & 0.799644
% & 19.2446 & 16.4392 & 0.8542 & 0.801992 \\

% Gemma3-27B
% & 21.0019 & 36.0892 & 0.8128 & 0.736109
% & 11.6991 & 11.3138 & 0.8243 & 0.768620 \\

% \bottomrule
% \end{tabular}
% }

% \caption{Translation performance comparison
% of nine models under the DI setting in both LZH-X and X-LZH
% translation directions.}
% \label{tab:translation_results_di}
% \end{table*}

\begin{table*}[t]
\centering
\small
\setlength{\tabcolsep}{5pt}
\renewcommand{\arraystretch}{1.2}

\resizebox{\textwidth}{!}{
\begin{tabular}{lcccccccc}
\toprule

\multicolumn{1}{c}{\multirow{2}{*}{Model(DT)}}
& \multicolumn{4}{c}{LZH-X}
& \multicolumn{4}{c}{X-LZH} \\

\cmidrule(lr){2-5}
\cmidrule(lr){6-9}

& BLEU & ChrF++ & BERTScore & COMET
& BLEU & ChrF++ & BERTScore & COMET \\

\midrule

Qwen3-8B
& 16.4975 & 32.4854 & 0.8008 & 0.711960
& 12.2196 & 11.5365 & 0.8101 & 0.759374 \\

Qwen3.8-27B
& 26.8670 & 41.7430 & 0.8374 & 0.778124
& 20.5774 & 16.8792 & 0.8443 & 0.788221 \\

Seed-X-PPO
& 16.0290 & 41.8972 & 0.8083 & 0.728496
& 6.8269 & 8.9710 & 0.7342 & 0.750249 \\

Seed-X-Instruct
& 14.6113 & 31.3028 & 0.7834 & 0.730740
& 5.9654 & 9.5262 & 0.7247 & 0.720437 \\

Qwen3.7
& 26.2831 & 41.1017 & 0.8373 & 0.777979
& 23.9789 & 19.5319 & 0.8444 & 0.789696 \\

Deepseek-flash
& 31.5811 & 45.2177 & 0.8477 & 0.780763
& \textbf{25.9394} & \textbf{20.7149} & \textbf{0.8735} & \textbf{0.811545} \\

GPT5.5
& \textbf{36.6704} & \textbf{49.2389} & \textbf{0.8642} & \textbf{0.804275}
& 21.7916 & 17.7811 & 0.8659 & 0.807057 \\

Gemini3.1
& 31.7368 & 45.6854 & 0.8511 & 0.799644
& 19.2446 & 16.4392 & 0.8542 & 0.801992 \\

Gemma3-27B
& 21.0019 & 36.0892 & 0.8128 & 0.736109
& 11.6991 & 11.3138 & 0.8243 & 0.768620 \\

\bottomrule
\end{tabular}
}

\caption{Translation performance comparison
of nine models under the DT setting in both LZH-X and X-LZH
translation directions. Best results in bold.}
\label{tab:translation_results_di}
\end{table*}

% Required packages in the preamble:
% \usepackage{booktabs}
% \usepackage{multirow}
% \usepackage{graphicx}

% \begin{table*}[t]
% \centering
% \small
% \setlength{\tabcolsep}{5pt}
% \renewcommand{\arraystretch}{1.2}

% \resizebox{\textwidth}{!}{
% \begin{tabular}{lcccccccc}
% \toprule

% \multicolumn{1}{c}{\multirow{2}{*}{Model(HGB)}}
% & \multicolumn{4}{c}{LZH-X}
% & \multicolumn{4}{c}{X-LZH} \\

% \cmidrule(lr){2-5}
% \cmidrule(lr){6-9}

% & BLEU & ChrF++ & BERTScore & COMET
% & BLEU & ChrF++ & BERTScore & COMET \\

% \midrule

% Qwen3-8B
% & 12.2149 & 24.3329 & 0.7734 & 0.645202
% & 29.0430 & 23.8852 & 0.8632 & 0.814734 \\

% Qwen3.8-27B
% & 30.2943 & 45.6224 & 0.8533 & 0.804146
% & 37.9433 & 28.7248 & 0.8861 & 0.824925 \\

% Seed-X-PPO
% & 10.8524 & 33.0707 & 0.7287 & 0.679762
% & 7.1642 & 11.8485 & 0.6925 & 0.691651 \\

% Seed-X-Instruct
% & 10.7300 & 32.1508 & 0.7247 & 0.667093
% & 8.9870 & 13.6161 & 0.7075 & 0.693047 \\

% Qwen3.7
% & 29.7738 & 44.9143 & 0.8520 & 0.799462
% & 42.4629 & 32.8775 & 0.8873 & 0.829411 \\

% Deepseek-flash
% & 31.0056 & 45.2062 & 0.8462 & 0.782737
% & 41.0149 & 31.6475 & 0.8963 & 0.835904 \\

% GPT5.5
% & 40.1176 & 53.5189 & 0.8764 & 0.835924
% & 34.9424 & 27.2901 & 0.9102 & 0.853857 \\

% Gemini3.1
% & 32.8266 & 47.7065 & 0.8589 & 0.821210
% & 35.8342 & 28.6618 & 0.8896 & 0.846840 \\

% Gemma3-27B
% & 25.7769 & 41.0833 & 0.8344 & 0.775935
% & 18.3080 & 16.1042 & 0.8539 & 0.800608 \\

% \bottomrule
% \end{tabular}
% }

% \caption{Translation performance comparison
% of nine models under the HGB setting in both
% LZH-X and X-LZH translation directions.}

% \label{tab:translation_results_hgb}

% \end{table*}
\begin{table*}[t]
\centering
\small
\setlength{\tabcolsep}{5pt}
\renewcommand{\arraystretch}{1.2}

\resizebox{\textwidth}{!}{
\begin{tabular}{lcccccccc}
\toprule

\multicolumn{1}{c}{\multirow{2}{*}{Model(HGB)}}
& \multicolumn{4}{c}{LZH-X}
& \multicolumn{4}{c}{X-LZH} \\

\cmidrule(lr){2-5}
\cmidrule(lr){6-9}

& BLEU & ChrF++ & BERTScore & COMET
& BLEU & ChrF++ & BERTScore & COMET \\

\midrule

Qwen3-8B
& 12.2149 & 24.3329 & 0.7734 & 0.645202
& 29.0430 & 23.8852 & 0.8632 & 0.814734 \\

Qwen3.8-27B
& 30.2943 & 45.6224 & 0.8533 & 0.804146
& 37.9433 & 28.7248 & 0.8861 & 0.824925 \\

Seed-X-PPO
& 10.8524 & 33.0707 & 0.7287 & 0.679762
& 7.1642 & 11.8485 & 0.6925 & 0.691651 \\

Seed-X-Instruct
& 10.7300 & 32.1508 & 0.7247 & 0.667093
& 8.9870 & 13.6161 & 0.7075 & 0.693047 \\

Qwen3.7
& 29.7738 & 44.9143 & 0.8520 & 0.799462
& \textbf{42.4629} & \textbf{32.8775} & 0.8873 & 0.829411 \\

Deepseek-flash
& 31.0056 & 45.2062 & 0.8462 & 0.782737
& 41.0149 & 31.6475 & 0.8963 & 0.835904 \\

GPT5.5
& \textbf{40.1176} & \textbf{53.5189} & \textbf{0.8764} & \textbf{0.835924}
& 34.9424 & 27.2901 & \textbf{0.9102} & \textbf{0.853857} \\

Gemini3.1
& 32.8266 & 47.7065 & 0.8589 & 0.821210
& 35.8342 & 28.6618 & 0.8896 & 0.846840 \\

Gemma3-27B
& 25.7769 & 41.0833 & 0.8344 & 0.775935
& 18.3080 & 16.1042 & 0.8539 & 0.800608 \\

\bottomrule
\end{tabular}
}

\caption{Translation performance comparison
of nine models under the HGB setting in both
LZH-X and X-LZH translation directions. Best results in bold.}

\label{tab:translation_results_hgb}

\end{table*}

\begin{figure*}[!t]
    \centering
    \setlength{\tabcolsep}{3pt}

    \begin{tabular}{@{}cc@{}}

    \includegraphics[width=0.47\textwidth,
        height=0.18\textheight,
        keepaspectratio]{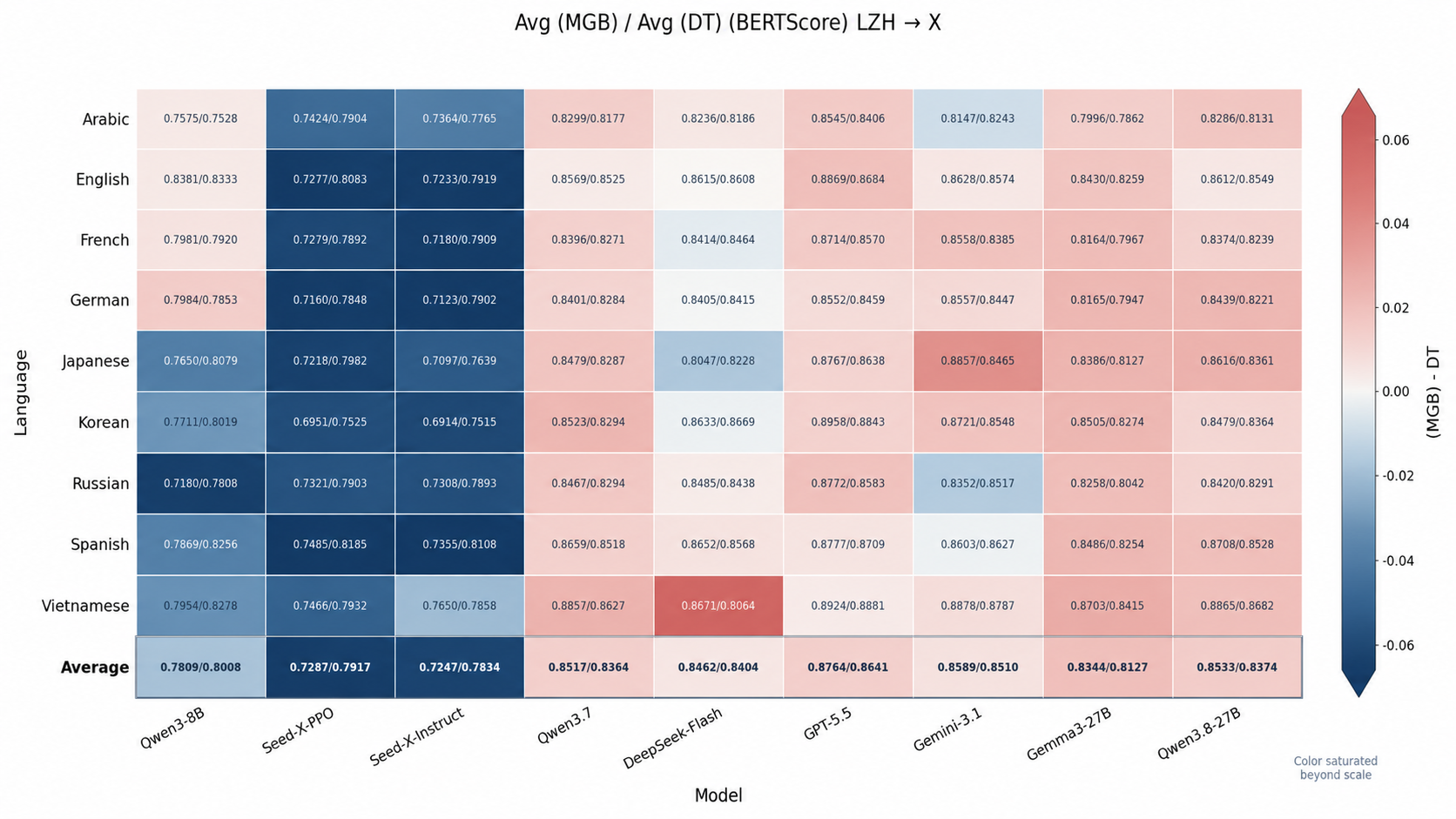} &
    \includegraphics[width=0.47\textwidth,
        height=0.18\textheight,
        keepaspectratio]{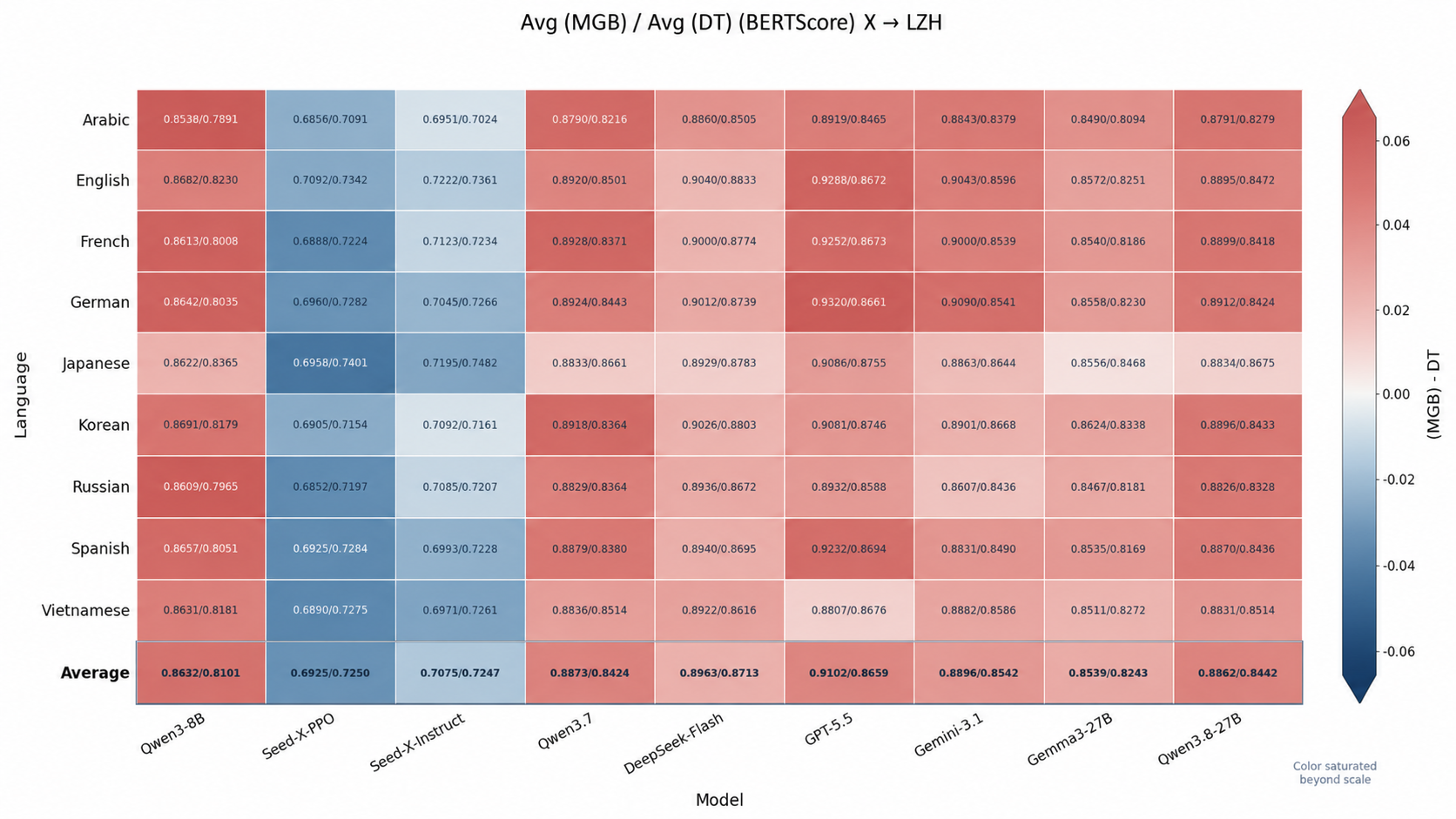} \\[3pt]
\multicolumn{2}{c}{\small (a) BERTScore comparison} \\[6pt]

    \includegraphics[width=0.47\textwidth,
        height=0.18\textheight,
        keepaspectratio]{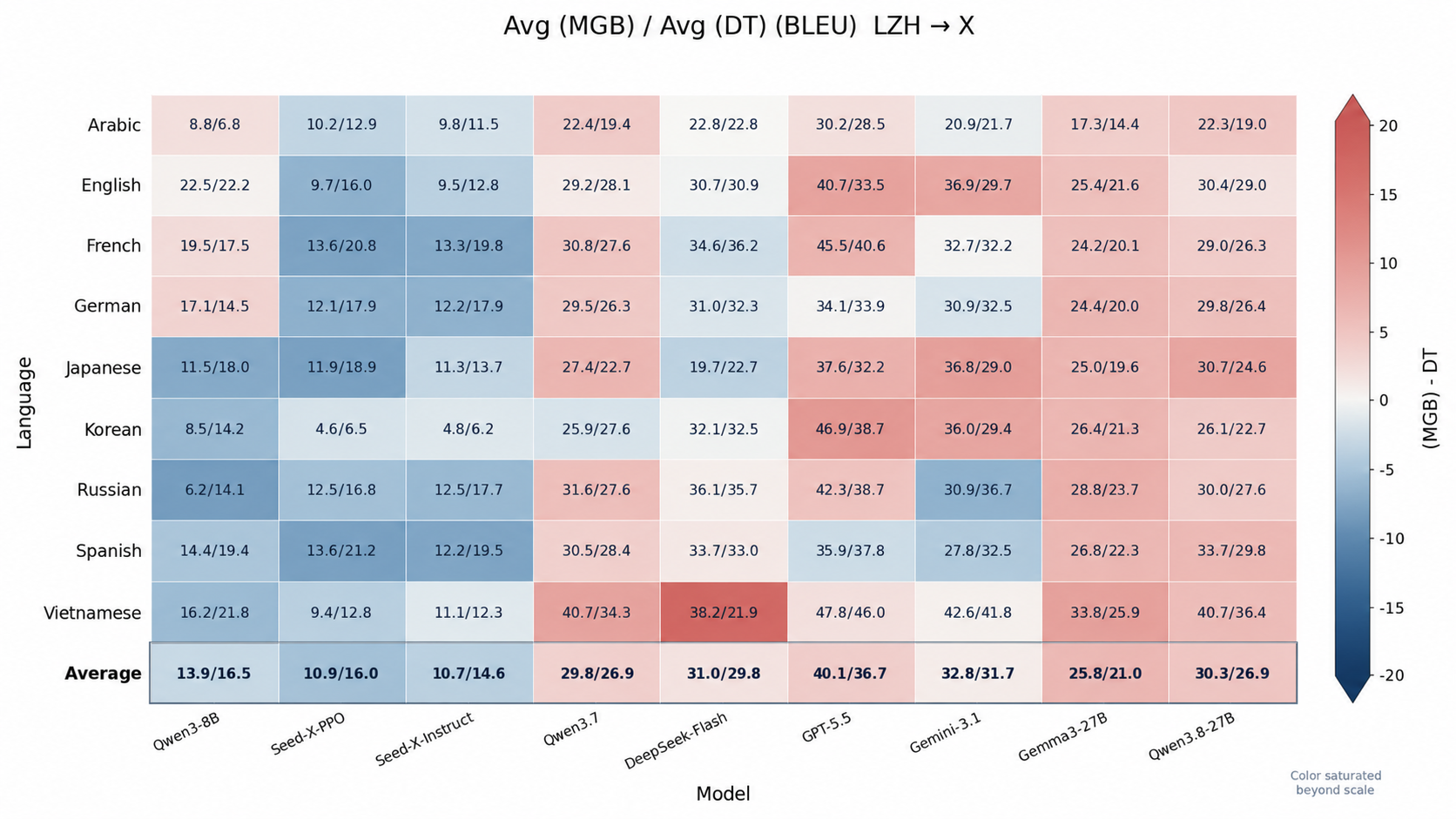} &
    \includegraphics[width=0.47\textwidth,
        height=0.18\textheight,
        keepaspectratio]{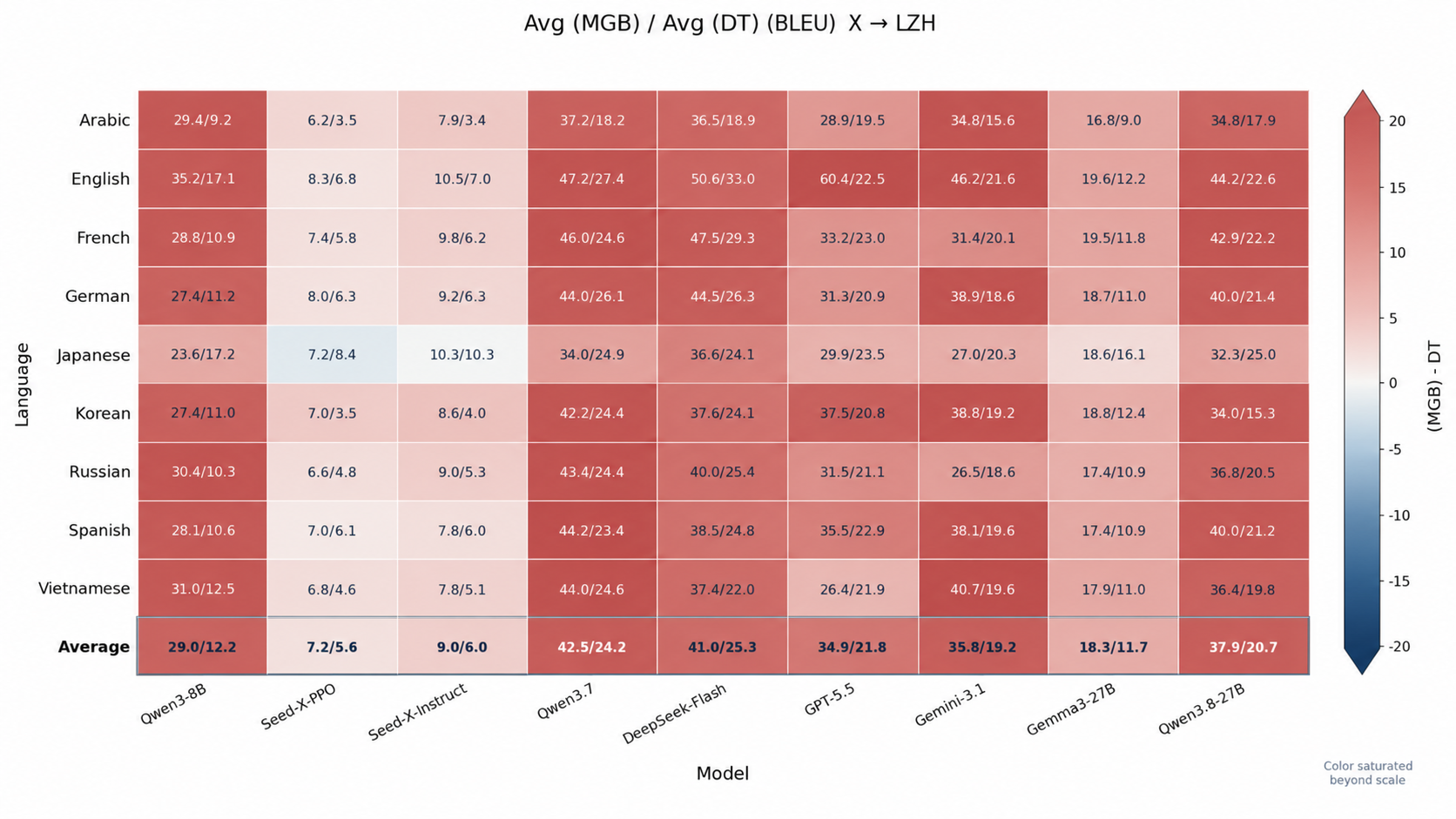} \\[3pt]
\multicolumn{2}{c}{\small (b) BLEU comparison} \\[6pt]

    \includegraphics[width=0.47\textwidth,
        height=0.18\textheight,
        keepaspectratio]{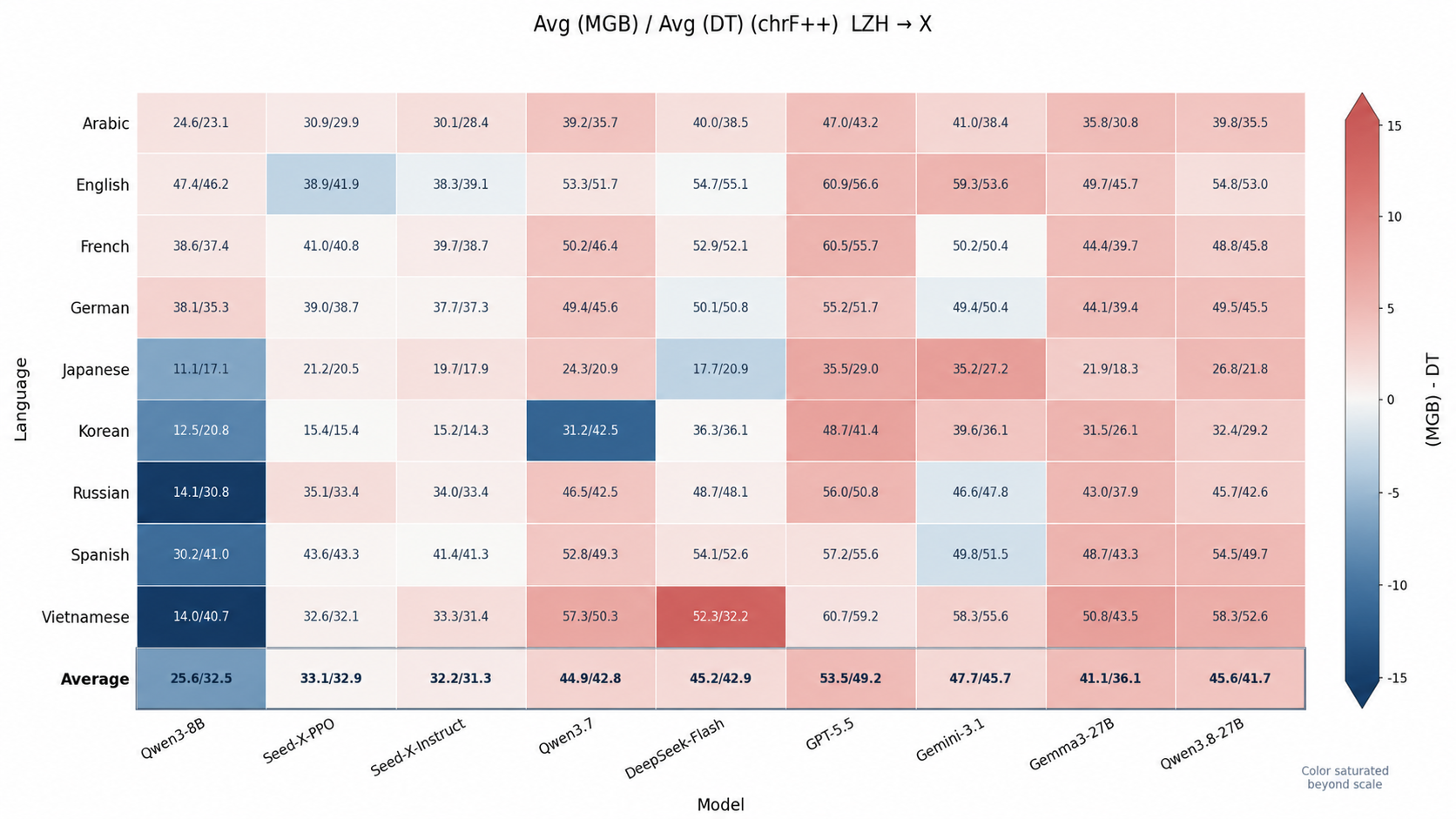} &
    \includegraphics[width=0.47\textwidth,
        height=0.18\textheight,
        keepaspectratio]{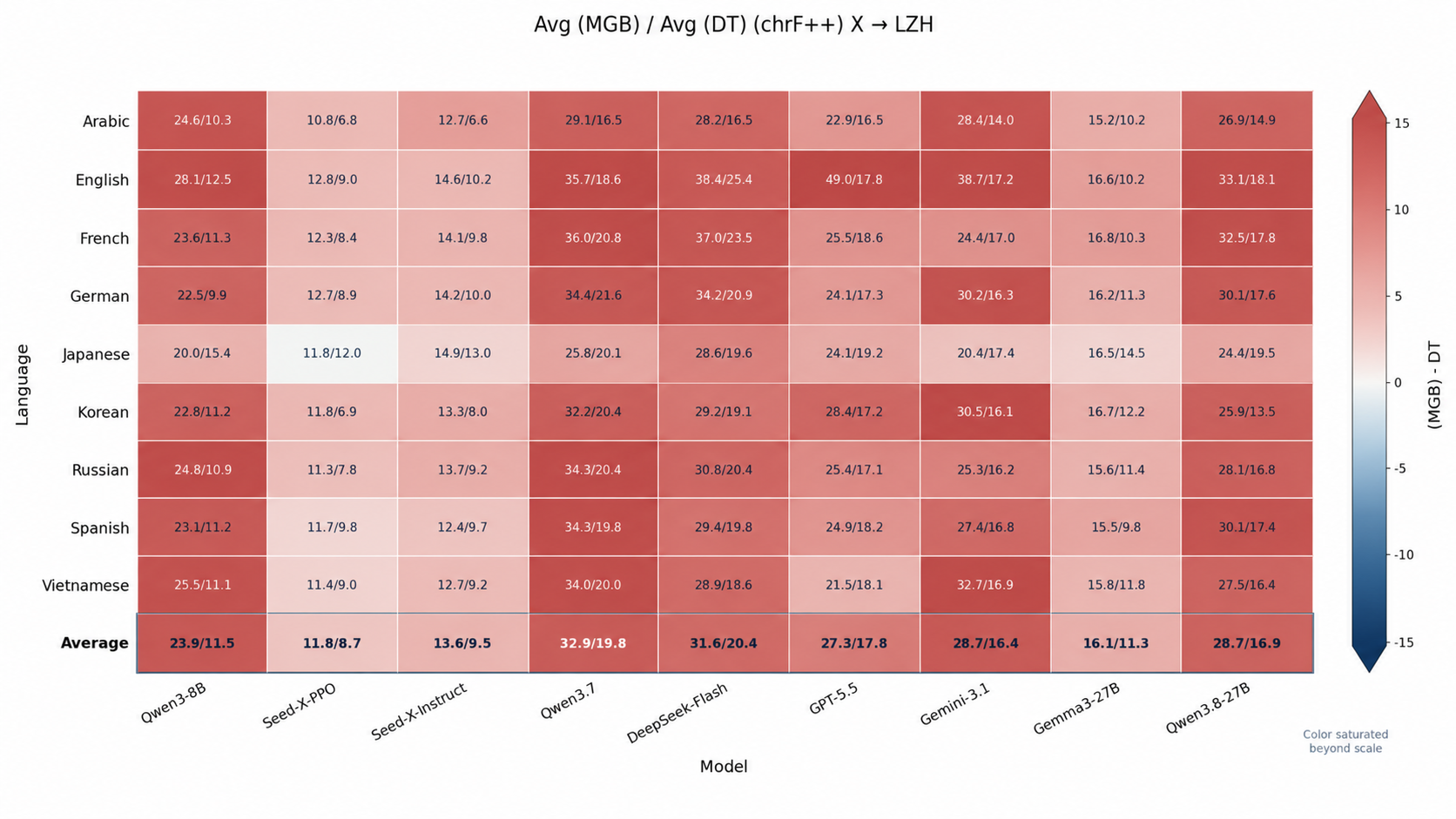} \\[3pt]
\multicolumn{2}{c}{\small (c) chrF++ comparison} \\[6pt]

        \includegraphics[width=0.47\textwidth,
        height=0.18\textheight,
        keepaspectratio]{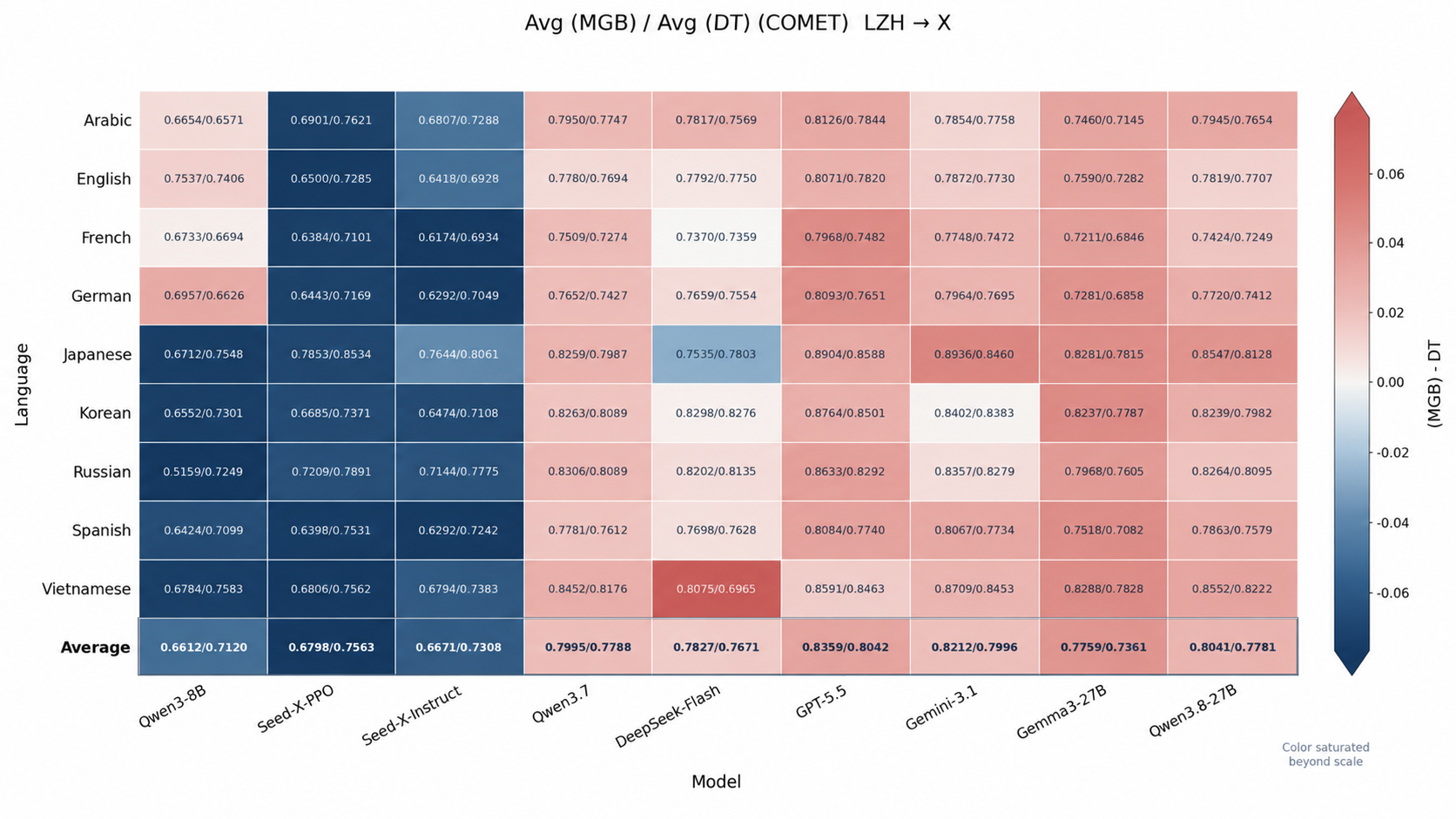} &
    \includegraphics[width=0.47\textwidth,
        height=0.18\textheight,
        keepaspectratio]{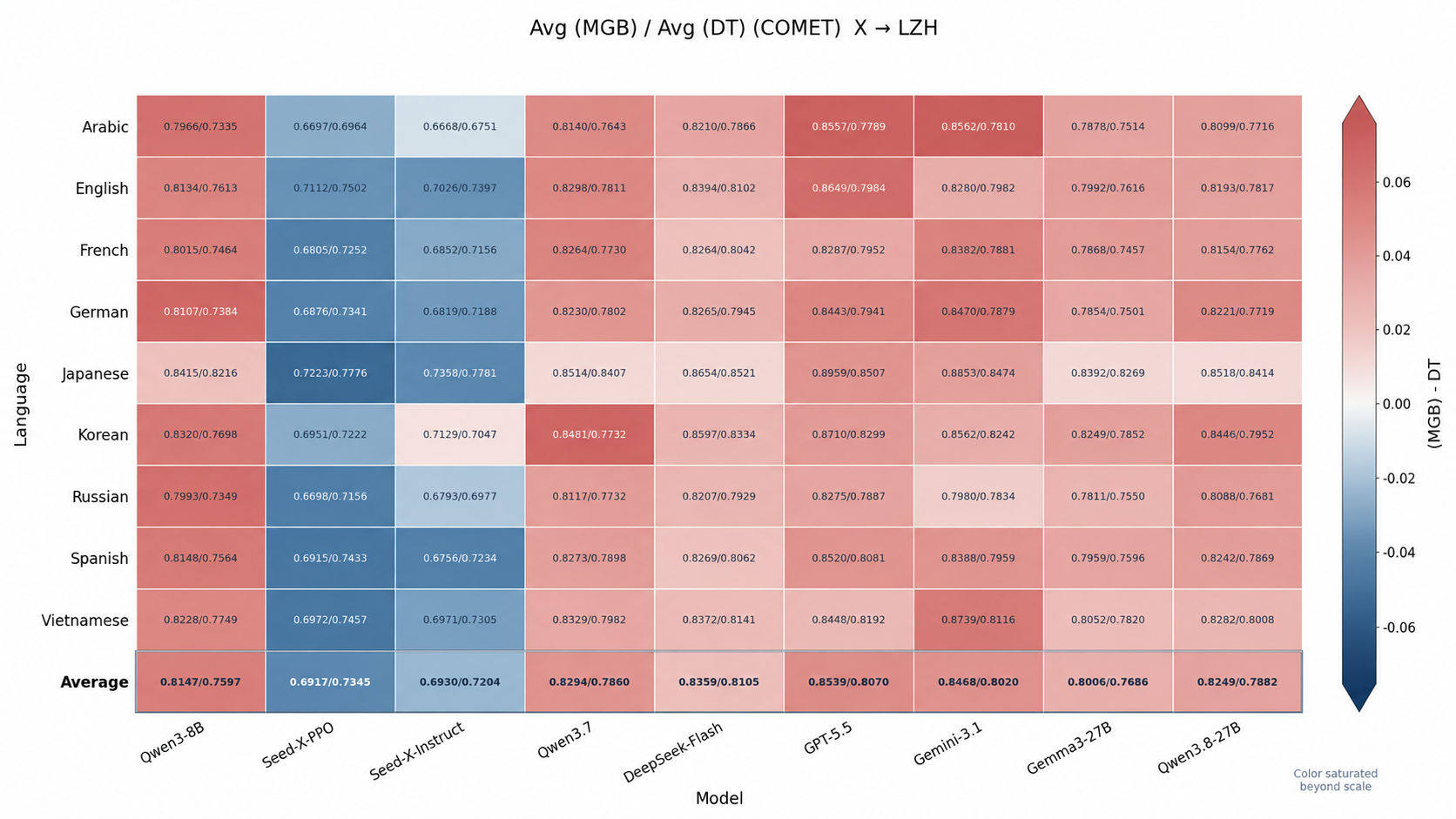} \\[3pt]
\multicolumn{2}{c}{\small (d) COMET comparison} \\[6pt]

    \end{tabular}

    \caption{Heatmap results comparing translation
    performance across different evaluation metrics
    and translation directions.}
    \label{fig:HeatMap}
\end{figure*}
Table~\ref{tab:translation_results_di} and Table~\ref{tab:translation_results_hgb} and Figure~\ref{fig:HeatMap} illustrate model performance in multilingual translation across BERTScore, BLEU, chrF++, and COMET. While overall metric distributions are consistent, pronounced differences exist across models, languages, and translation directions. Vertical heatmap stratification indicates that a model's intrinsic cross-lingual capability dictates its baseline performance, whereas local fluctuations reflect the effects of specific language combinations.

Semantic Performance (BERTScore): Figure (a) shows significant inter-model variance in the left translation direction, contrasted by uniformly higher scores (warmer colors) in the right. This asymmetry demonstrates that a model's ability to preserve semantics is highly direction-dependent and influenced by language structure and training data distribution.

Lexical \& Character Matching (BLEU \& chrF++): Figures (b) and (c) exhibit similar directional asymmetry. However, local discrepancies between the two metrics highlight differing sensitivities to morphological and expression changes, proving that a single lexical metric cannot fully capture translation quality.

Comprehensive Semantics (COMET): Figure (d) aligns closely with BERTScore, confirming that inter-model differences stem from deep semantic representation rather than surface-level matching alone. Language-specific fluctuations further indicate that model generalization is constrained by resource availability, structural complexity, and pre-training data coverage.

Conclusion: Multilingual translation quality is jointly shaped by model capability, directional asymmetry, and linguistic features. Because semantic and lexical metrics capture distinct dimensions of quality, a reliable model evaluation requires a comprehensive, multi-dimensional approach combining both metrics across different translation directions.
\subsection{Preference Analysis}
\begin{figure*}[!t]
    \centering
    \setlength{\tabcolsep}{3pt}

    \begin{tabular}{@{}cc@{}}

    \includegraphics[width=0.47\textwidth,
        height=0.18\textheight,
        keepaspectratio]{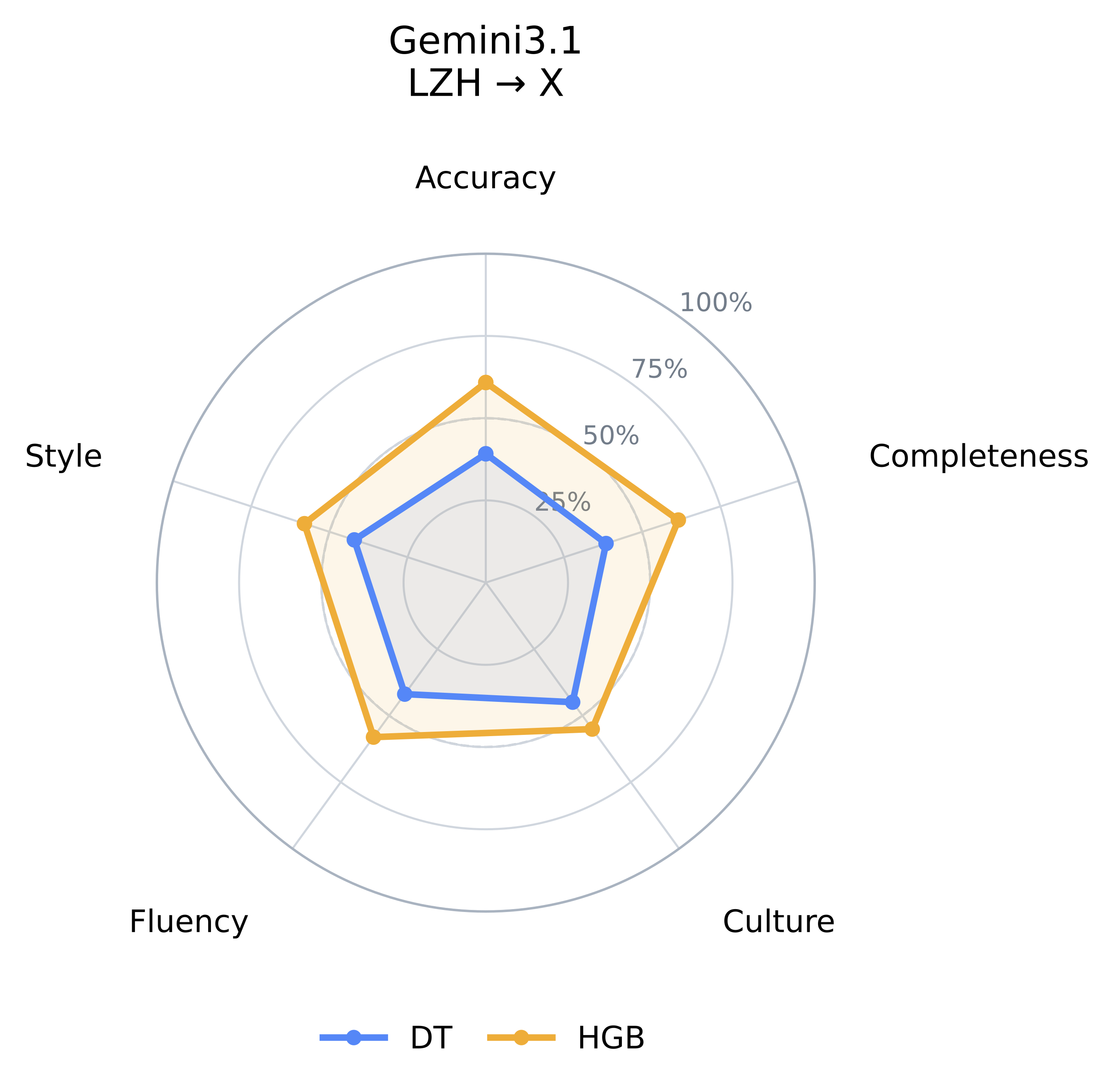} &
    \includegraphics[width=0.47\textwidth,
        height=0.18\textheight,
        keepaspectratio]{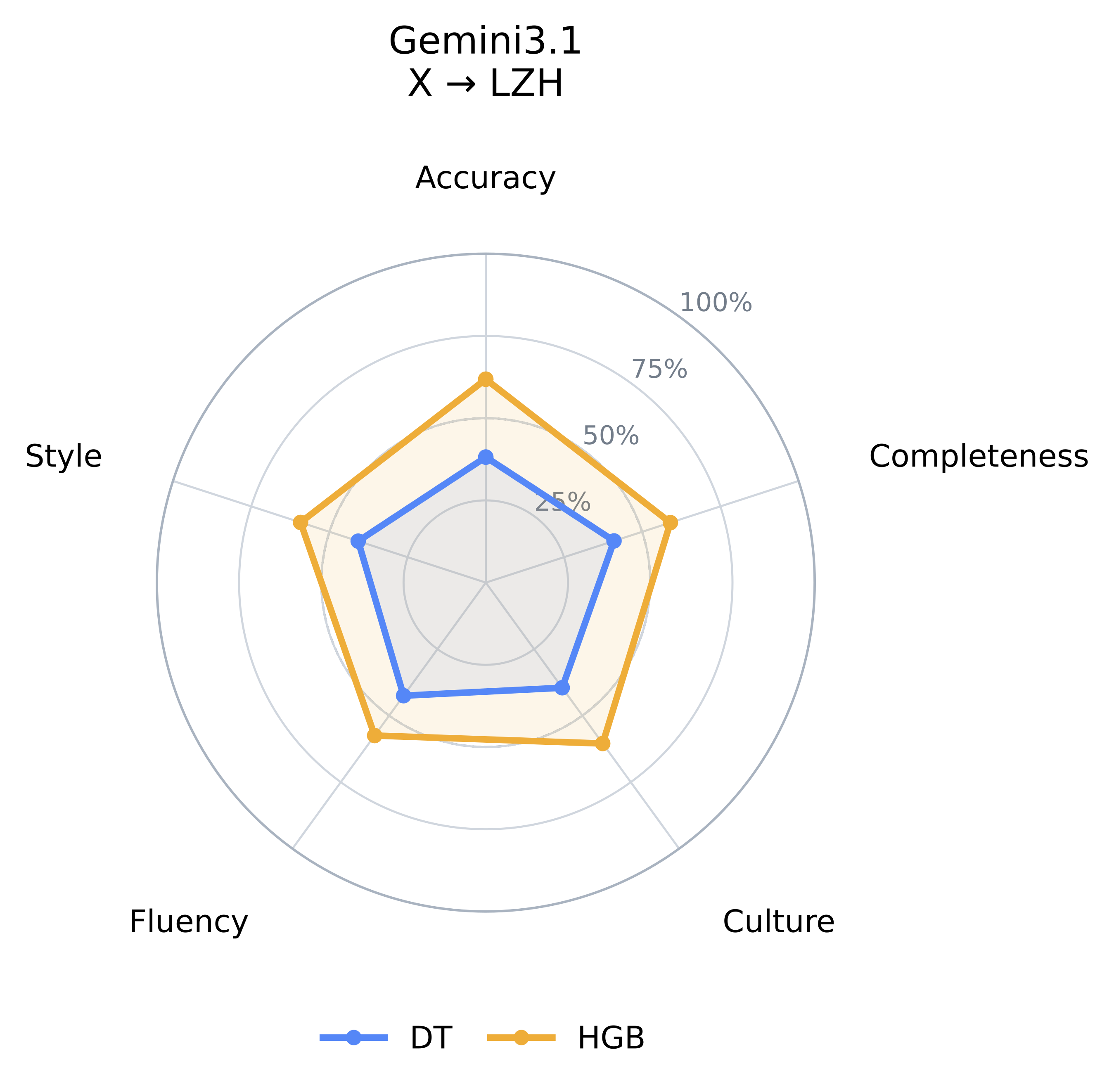} \\[3pt]
\multicolumn{2}{c}{\small (a) Gemini3.1 comparison} \\[6pt]
    \includegraphics[width=0.47\textwidth,
        height=0.18\textheight,
        keepaspectratio]{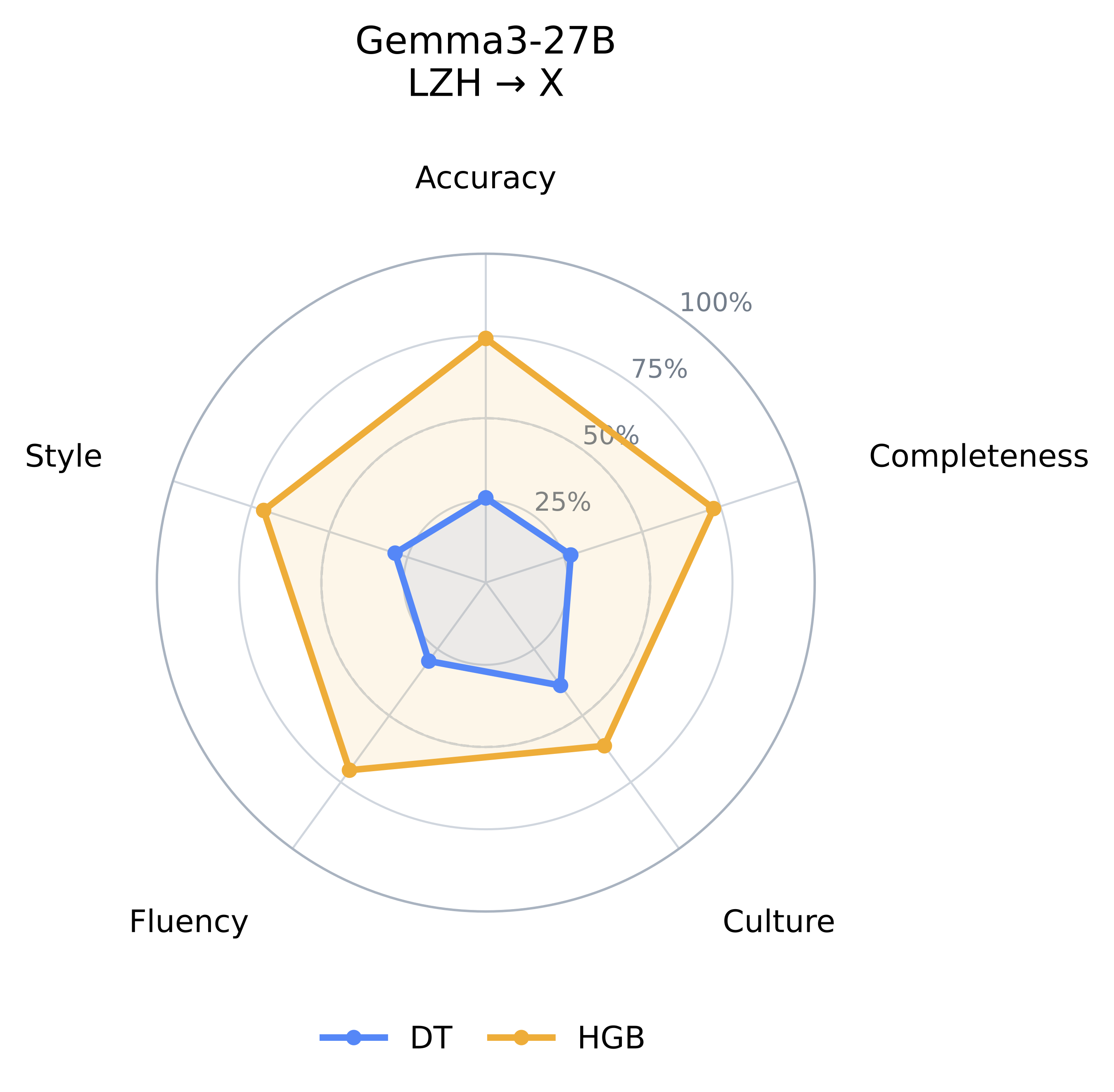} &
    \includegraphics[width=0.47\textwidth,
        height=0.18\textheight,
        keepaspectratio]{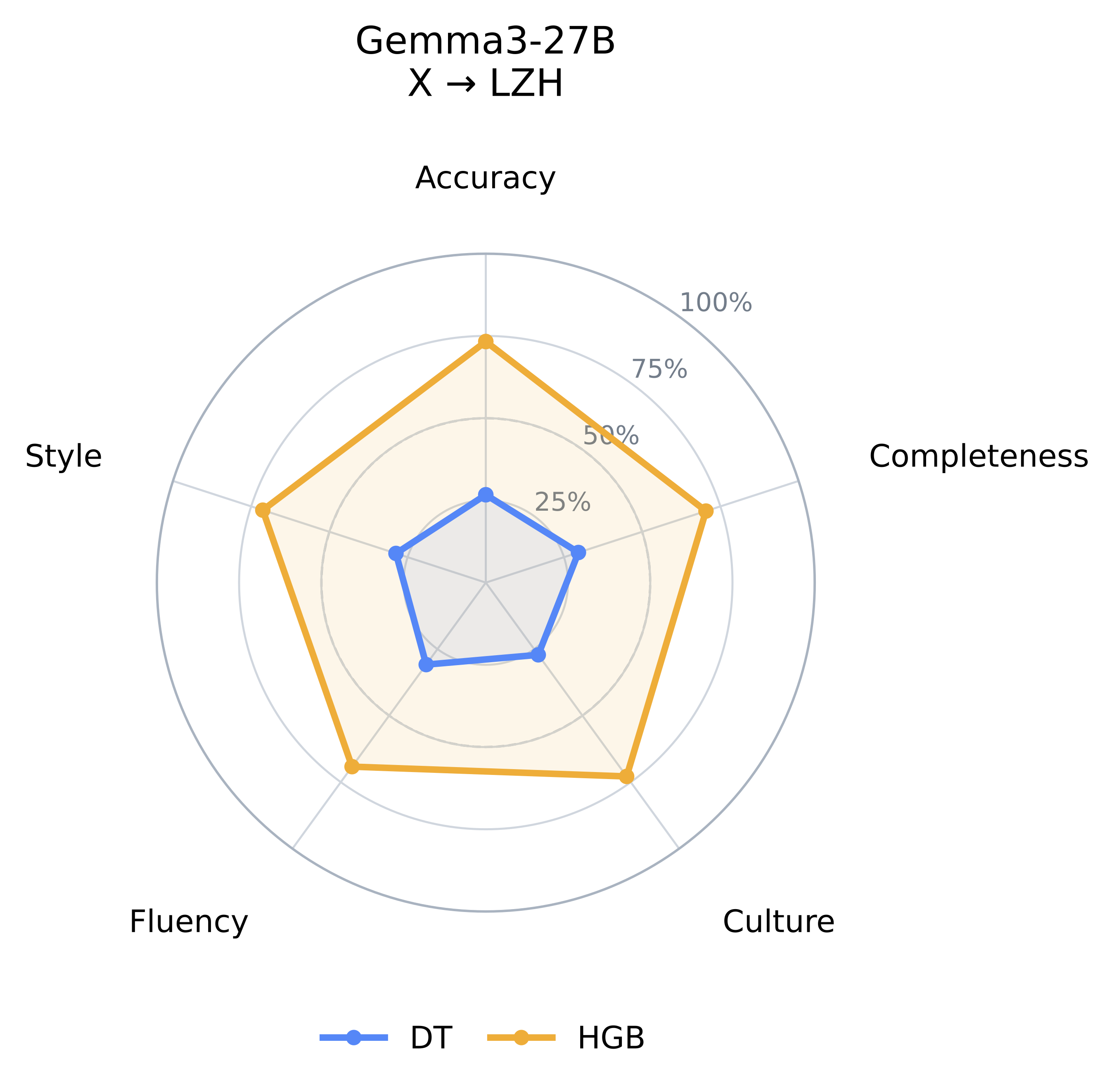} \\[3pt]
\multicolumn{2}{c}{\small (b) Gemma3-27B comparison} \\[6pt]
    \includegraphics[width=0.47\textwidth,
        height=0.18\textheight,
        keepaspectratio]{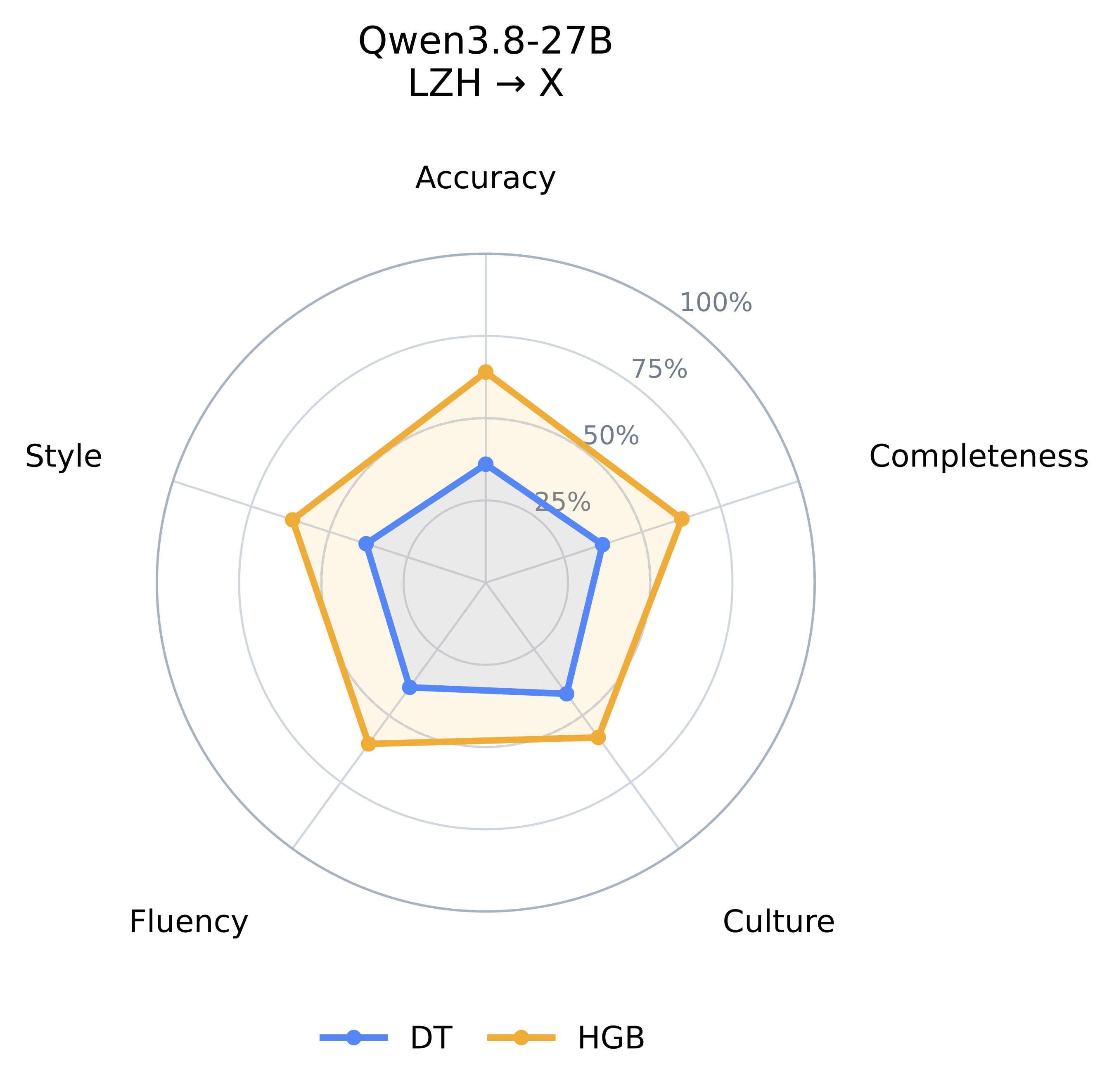} &
    \includegraphics[width=0.47\textwidth,
        height=0.18\textheight,
        keepaspectratio]{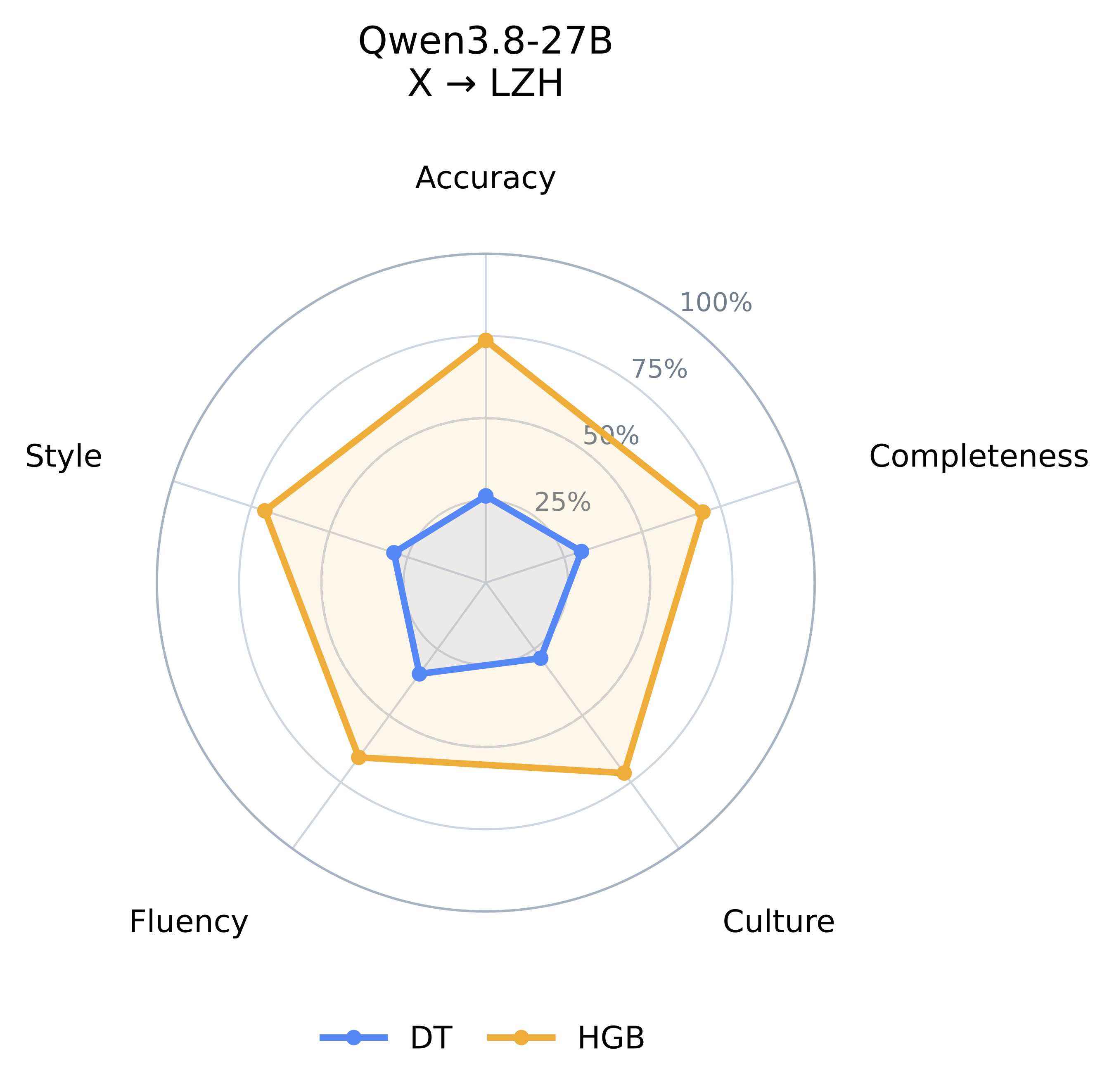} \\[3pt]
\multicolumn{2}{c}{\small (c) Qwen3.8-27B comparison} \\[6pt]
    \end{tabular}

    \caption{Direct Translation (DT) and Human-Gloss Bridged performance (HGB)
    across different models in LZH--X and X--LZH
    translation directions.}
    \label{fig:Radar}
\end{figure*}

Figure~\ref{fig:Radar} presents a multidimensional native-speaker expert evaluation (Accuracy, Completeness, Culture, Fluency, and Style) of three LLMs in bidirectional Classical Chinese translation. Across all models, the semantic bridge strategy (HGB) consistently outperforms the baseline translation (DT), delivering comprehensive improvements across all five dimensions.

LZH→X (Classical to Other): Semantic bridges effectively unpack highly condensed Classical Chinese. While Gemini 3.1 improves steadily from a strong baseline, Gemma3-27B and Qwen experience substantial leaps—particularly in Accuracy, Fluency, and Style—proving the strategy preserves both core semantics and cultural nuances.
X→LZH (Other to Classical): Generating Classical Chinese demands strict adherence to unique syntactic and cultural norms. Baseline models (especially Gemma3-27B and Qwen) struggle significantly here, but HGB provides structural semantic guidance, driving dramatic gains in Accuracy, Culture, and Style.
Impact by Dimension: Improvements in Accuracy and Completeness indicate reduced information loss during cross-lingual transfer. Gains in Culture and Style successfully balance semantic fidelity with target-language norms, while higher Fluency ensures natural readability.
Model Dependency \& Applicability: The strategy acts as a powerful compensatory mechanism. Models with weaker baselines (Gemma3-27B, Qwen) see massive performance boosts, whereas stronger models (Gemini 3.1) yield more moderate enhancements. Despite varying initial capabilities, the universal improvements demonstrate robust cross-model applicability.

Ultimately, these human evaluations corroborate automated metrics, confirming that semantic bridges optimize nuanced cultural and stylistic expressions that automated metrics often miss. By mitigating profound semantic and structural gaps, intermediate semantic representations offer a highly effective optimization for low-resource classical language translation.

\subsection{Semantic Comparison Analysis from Different Sources}

\begin{figure*}[!t]
    \centering
    \setlength{\tabcolsep}{3pt}

    \begin{tabular}{@{}cc@{}}

    \includegraphics[width=0.47\textwidth,
        height=0.18\textheight,
        keepaspectratio]{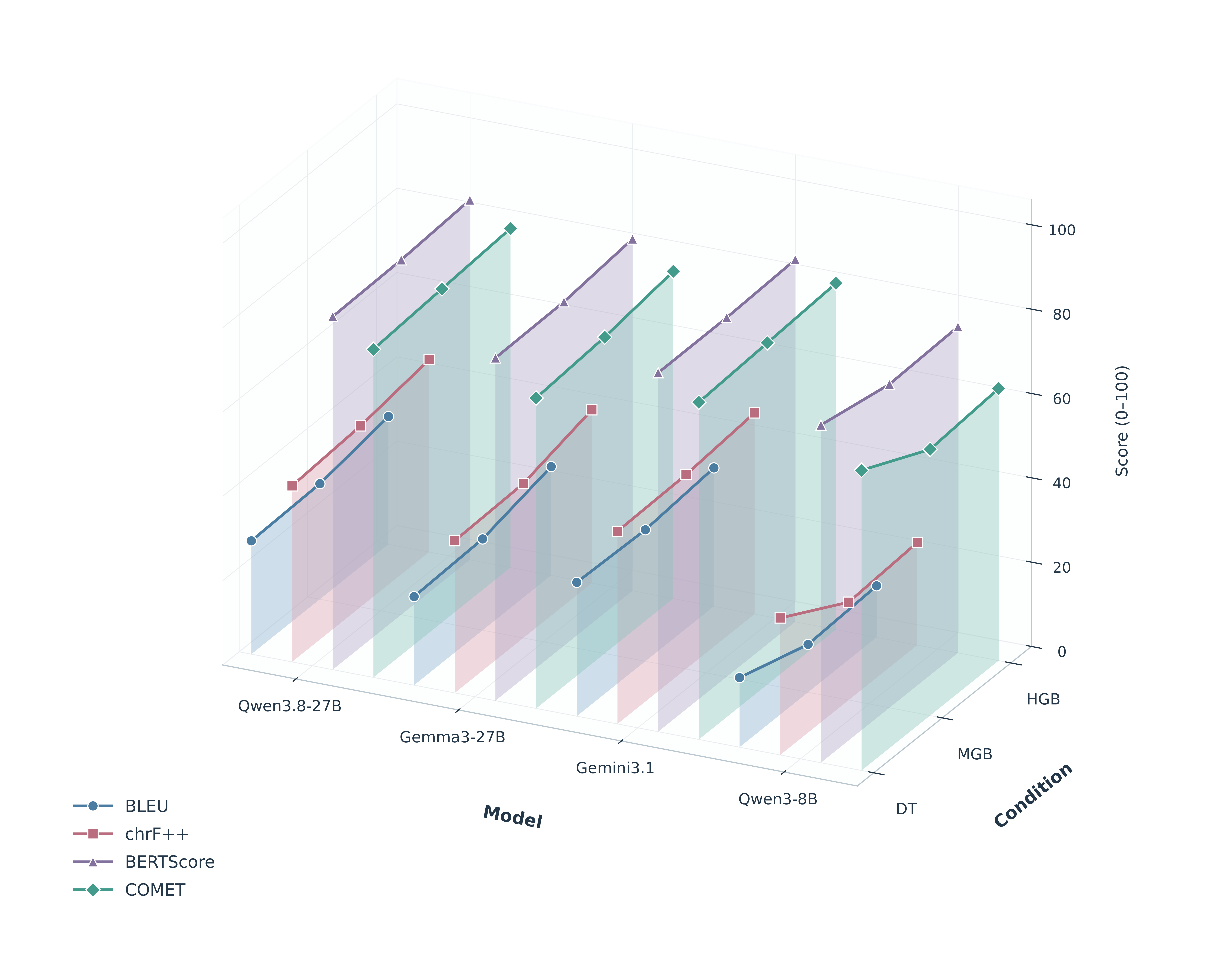} &
    \includegraphics[width=0.47\textwidth,
        height=0.18\textheight,
        keepaspectratio]{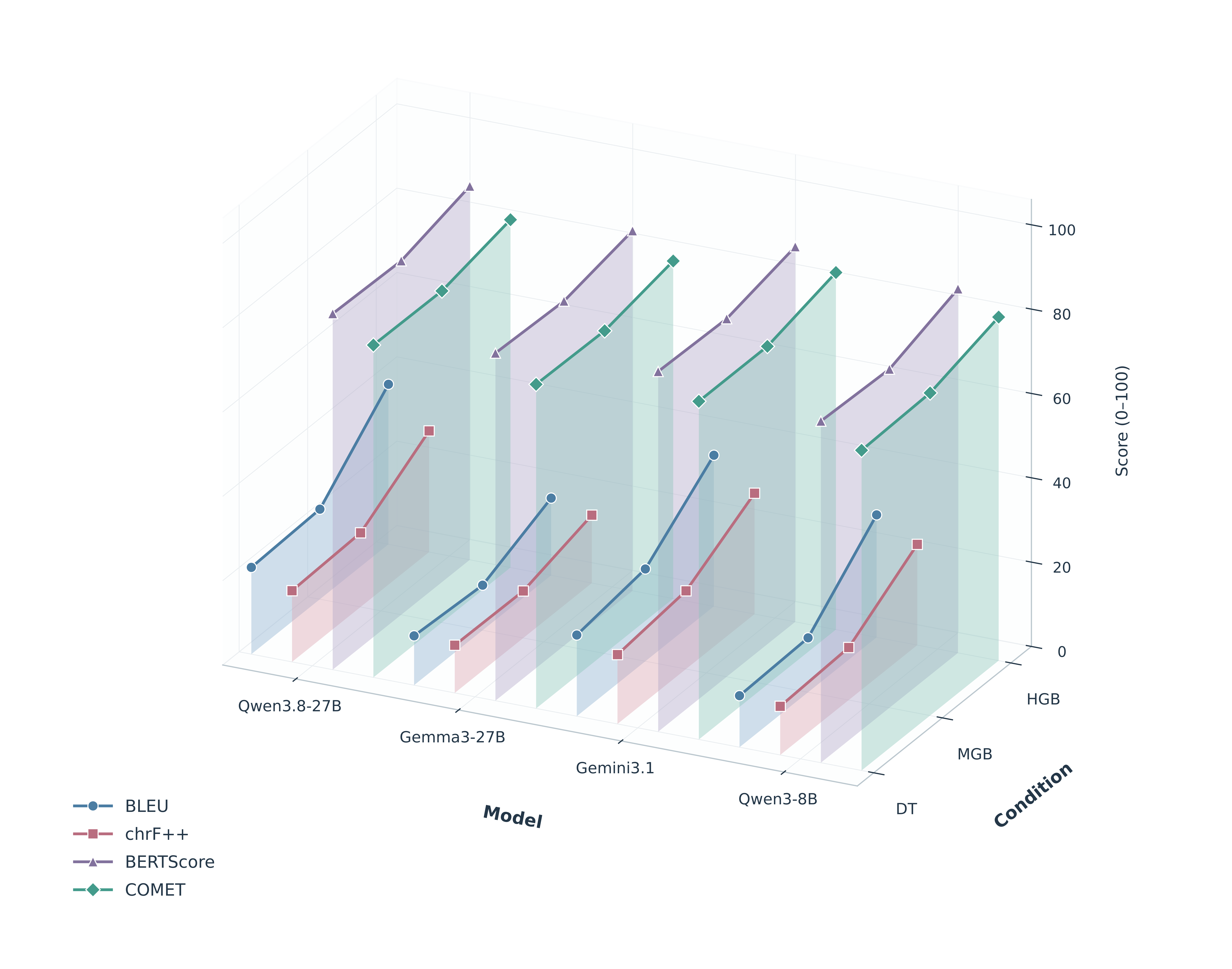} \\[3pt]
\multicolumn{2}{c}{\small(b) X--LZH} \\[6pt]
   
    \end{tabular}

    \caption{Comparison chart of DT/HGB/MGB.}
    \label{fig:wave}
\end{figure*}

Figure~\ref{fig:wave} evaluates the performance of four LLMs in bidirectional Classical Chinese translation (LZH-X and X-LZH) using three setups: basic translation (DT) and semantic bridge-enhanced (HGB, MGB) across BLEU, chrF++, BERTScore, and COMET. Overall, semantic bridges consistently improve translation quality across all metrics and directions, with HGB performing best, followed by MGB, and DT ranking lowest.

LZH-X (Classical to Other): Classical Chinese’s concise, context-heavy nature often causes semantic loss during direct translation. By providing explicit intermediate representations, both MGB and HGB significantly improve lexical matching (BLEU/chrF++) and semantic integrity (BERTScore/COMET), with MGB yielding the highest gains.
X-LZH (Other to Classical): Baseline scores (DT) are inherently lower due to the difficulty of generating complex, condensed Classical Chinese syntax. Semantic bridges markedly boost performance by providing a clear semantic structure, facilitating both accurate source comprehension and target language expression.
Cross-Model Applicability: While larger models achieve higher baselines (with Qwen3-8B performing weakest), the semantic bridge strategy universally benefits all models. It is particularly effective at compensating for the cross-lingual deficiencies of weaker models, though specific gains vary by model architecture and pre-trained knowledge.

Ultimately, explicit intermediate semantic representations—especially the HGB strategy—effectively bridge the significant structural and semantic gaps in bidirectional Classical Chinese translation. The consistent improvements across lexical and semantic metrics validate this approach as a robust optimization for complex, low-resource languages.

\subsection{Model details}
\label{app}
Table~\ref{app:model_details} summarizes the model versions, inference configurations, and source platforms used in our experiments. All models are configured with a temperature of 0.0 and a maximum output length of 256 tokens to reduce sampling variability and maintain a consistent generation budget. The context-window limits range from 32K to 1M tokens, reflecting model-specific capacities. The checkpoint or API identifiers and corresponding platforms are provided to document the experimental setup and support reproducibility.

\begin{table*}[t]
\centering
\footnotesize
\setlength{\tabcolsep}{3pt}
\renewcommand{\arraystretch}{1.3}
\urlstyle{same}

\begin{tabular}{@{}
p{0.125\textwidth}
p{0.205\textwidth}
p{0.115\textwidth}
p{0.095\textwidth}
p{0.11\textwidth}
p{0.255\textwidth}
@{}}

\toprule[1.2pt]
\textbf{Model} &
\textbf{Model Version} &
\textbf{Temperature} &
\shortstack[l]{\textbf{Context}\\\textbf{Window}} &
\shortstack[l]{\textbf{Max Output}\\\textbf{Tokens}} &
\textbf{Source / Platform} \\
\midrule[0.5pt]

Qwen3-8B &
\path{Qwen/Qwen3-8B} &
0.0 & 32K & 256 &
\url{https://huggingface.co/Qwen/Qwen3-8B} \\

Qwen3.8-27B &
\path{Qwen/Qwen3.8-27B} &
0.0 & 256K & 256 &
\url{https://huggingface.co/Qwen/Qwen3.8-27B} \\

Seed-X-PPO &
\path{ByteDance-Seed/Seed-X-PPO-7B} &
0.0 & 32K & 256 &
\url{https://huggingface.co/ByteDance-Seed/Seed-X-PPO-7B} \\

Seed-X-Instruct &
\path{ByteDance-Seed/Seed-X-Instruct-7B} &
0.0 & 32K & 256 &
\url{https://huggingface.co/ByteDance-Seed/Seed-X-Instruct-7B} \\

Gemma3-27B &
\path{google/gemma-3-27b-it} &
0.0 & 128K & 256 &
\url{https://huggingface.co/google/gemma-3-27b-it} \\

DeepSeek-Flash &
\path{deepseek-ai/DeepSeek-V4.1-Flash} &
0.0 & 1M & 256 &
\url{https://platform.deepseek.com/} \\

GPT5.5 &
\path{gpt-5.5-2026-04-23} &
0.0 & 1M & 256 &
\url{https://platform.openai.com/} \\

Gemini3.1 &
\path{gemini-3.1-flash-lite} &
0.0 & 1M & 256 &
\url{https://aistudio.google.com/} \\

Qwen3.7 &
\path{qwen3.7-flash} &
0.0 & 1M & 256 &
\url{https://qwen.ai/apiplatform} \\

\bottomrule[1.2pt]

\end{tabular}

\caption{Hugging Face identifiers and inference configurations of the models used in our experiments.}
\label{app:model_details}

\end{table*}

\end{document}